%% file: main.tex
\documentclass{article}

\PassOptionsToPackage{numbers, compress}{natbib}
\PassOptionsToPackage{table}{xcolor}
\usepackage[preprint]{neurips_2026}

\usepackage[utf8]{inputenc} 
\usepackage[T1]{fontenc}    
\usepackage[colorlinks=true, linkcolor=blue]{hyperref}       
\usepackage{url}            
\usepackage{booktabs}       
\usepackage{amsfonts}       
\usepackage{nicefrac}       
\usepackage{microtype}      
\usepackage{xcolor}         
\usepackage{graphicx}       
\usepackage{multicol}
\usepackage{makecell}
\usepackage{multirow}
\usepackage[font=small,skip=3pt]{caption}
\usepackage{algorithm}
\usepackage[noend]{algpseudocode}
\usepackage{titletoc} 
\usepackage[normalem]{ulem}  

\usepackage{orcidlink}

\input{shortcuts}

\title{TRON: \textbf{T}racing \textbf{R}ays to \textbf{O}rchestrate a \textbf{N}eural Renderer for 3D
Gaussian Reconstructions}

\author{%
  \hspace{-4mm}Or Perel\orcidlink{0000-0001-6478-1422}$^{1,2,3}$ \quad 
  Hassan Abu Alhaija\orcidlink{0000-0001-6669-7072}$^{1}$ \quad 
  Zian Wang\orcidlink{0000-0003-4166-3807}$^{1,2,3}$ \quad 
  Jacob Munkberg\orcidlink{0009-0004-0451-7442}$^{1}$ \quad \vspace{8pt}\\
  \textbf{Matan Atzmon}\orcidlink{0009-0001-6998-1033}$^{1}$ \quad 
  \textbf{Sanja Fidler} \orcidlink{0000-0003-1040-3260}$^{1,2,3}$ \quad 
  \textbf{Maria Shugrina}\orcidlink{0000-0002-7583-6772}$^{1}$ \vspace{8pt} \\
  \small{\textsuperscript{1}NVIDIA \quad \textsuperscript{2}University of Toronto \quad \textsuperscript{3}Vector Institute } \vspace{1pt}\\
}

\begin{document}

\maketitle

\input{Sections/00_abstract}
\input{Sections/01_introduction}
\input{Sections/02_related}
\input{Sections/03_method_tracer}
\input{Sections/04_method_neural_renderer}
\input{Sections/05_evaluation}
\input{Sections/07_limitations_conclusion}

%
\bibliographystyle{plainnat}
\bibliography{main}

\newpage
\appendix

\section*{Appendix} 
\startcontents[appendix]
\printcontents[appendix]{}{1}{\setcounter{tocdepth}{2}}
\vspace{1em} 
\hrule 
\vspace{2em}

\newpage
\input{Sections/appendix}

\end{document}

%% file: shortcuts.tex
\usepackage{amsmath}          
\usepackage{xspace}           
\usepackage{url}              
\usepackage{xcolor}           

\usepackage{subcaption}       
\usepackage{caption}          
\usepackage{tikz}             
\usepackage{pgfplots}         
\usepackage{amssymb}

\usepackage[most]{tcolorbox}
\usepackage{enumitem}

\newcommand{\ourmethod}{TRON}

\newcommand{\tdgrt}{3DGRT}

\newcommand{\tronsynth}{TRON-Synth}

\newcommand{\dldvbenchmark}{TRON-DL3DV-33}

\providecommand{\best}[1]{\textbf{#1}}

\providecommand{\best}[1]{\textbf{#1}}

\newcommand{\bb}{\mathbf{b}}

\newcommand{\bd}{\mathbf{d}}

\newcommand{\bn}{\mathbf{n}}
\newcommand{\bo}{\mathbf{o}}

\newcommand{\bR}{\mathbf{R}}
\newcommand{\bs}{\mathbf{s}}

\newcommand{\bbeta}{\boldsymbol{\beta}}

\newcommand{\bmu}{\boldsymbol{\mu}}

\newcommand{\bomega}{\boldsymbol{\omega}}

\newcommand{\cG}{\mathcal{G}}

\usepackage{amsfonts}

\makeatletter
\DeclareRobustCommand\onedot{\futurelet\@let@token\@onedot}
\def\@onedot{\ifx\@let@token.\else.\null\fi\xspace}
\def\eg{e.g\onedot}

\makeatother

\definecolor{darkred}{rgb}{0.7,0.2,0.1}
\definecolor{darkgreen}{rgb}{0.2,0.7,0}
\definecolor{orange}{RGB}{255,127,0}
\definecolor{ourpurple}{RGB}{127,127,204}
\definecolor{ourteal}{RGB}{60,160,160}
\definecolor{palgreen}{RGB}{51,179,179}
\definecolor{magenta}{RGB}{199,21,133}
\definecolor{olive}{RGB}{100,150,85}



%% file: Sections/00_abstract.tex
\begin{abstract}

We introduce TRON, a rendering framework that combines 3D Gaussian ray tracing with neural rendering to enable realistic and controllable rendering of real-world 3D scenes under novel lighting, dynamic object motion, object insertion, and material editing.
Prior approaches that rely solely on physically based rendering (PBR) of Gaussian representations struggle to achieve realistic relighting due to imperfections in reconstructed geometry, material estimates, and light transport estimation. At the same time, neural rendering methods often lack an explicit scene representation, limiting their ability to support interactive editing with fine-grained manipulation.
TRON bridges these two paradigms. We use intrinsic decomposition priors from a learned inverse rendering model to regularize the material properties of a Gaussian field, and repurpose a ray tracer to provide radiometric guidance rather than final pixels. By treating this output as a structured 3D scaffold, we empower a lightweight neural renderer to bridge the domain gap between shading-model constrained estimates and photorealistic output.
Our key insight is that the combination of explicit 3D knowledge with robust material priors provides speed and controllability, while neural rendering enables the synthesis of photorealistic images.
To support real-world scenarios, we train our neural renderer with a multi-stage strategy consisting of large-scale pretraining and targeted fine-tuning on a newly constructed dataset of 2.1M rendered synthetic and real-world frames from 3D reconstructions.
TRON outperforms Gaussian-based relighting methods in realism, and prior neural renderers in editability and speed.
To the best of our knowledge, TRON is the first method to enable practical interactive applications in captured 3D environments, offering realistic appearance under dynamic geometric, lighting and material conditions.

\textbf{Keywords:} Neural Rendering, 3D Gaussian Ray Tracing, Inverse Rendering, Relighting

\end{abstract}

%% file: Sections/01_introduction.tex
\section{Introduction}
\label{sec:intro}

Recent advances have propelled multi-view reconstructions based on 3D 
Gaussian Splats (3DGS) \cite{kerbl3Dgaussians,wu20253dgut,3dgrt2024} toward interactive use cases, enabling elasto-dynamic simulation \cite{modi2024simplicits,xie2023physgaussian},
mechanical property prediction \cite{dagli2025vomp}, and painting
\cite{pandey2025painting} directly with 3D Gaussians.
Despite promising demonstrations~\cite{pandey2025realtimelive,modi2025lab}, broad adoption in interactive settings remains limited, as such applications require rendering under dynamic lighting and object motion. A key limitation of Gaussian representations is their reliance on \emph{baked shading}, where lighting effects (\eg, shadows and reflections) are encoded in a fixed scene radiance. 
Motivated to lift these barriers, we introduce {\ourmethod}, a novel framework that extends Gaussian ray tracing \cite{3dgrt2024} with explicit modeling of lighting effects coupled with a neural renderer, enabling realistic rendering under dynamic lighting, moving objects, and varying materials.

Our rendering solution considers the following key requirements: (i) direct 3D control with dynamic materials, illumination and object poses, (ii) interactive response to these dynamic edits (no pre-computation), and (iii) realistic rendering consistent with these controls. 
Recent approaches extending 3DGS to physically-based rendering (PBR) can enable some editing \cite{fan2025rng,R3DG2023,liang2023gs,du2024gsidilluminationdecompositiongaussian}, but
offer limited support of dynamic object motion, and often struggle to 
produce realistic results due to imperfect geometry, materials, and light transport computation in this under-constrained intrinsic decomposition problem. 
At the same time, emerging neural rendering approaches~\cite{DiffusionRenderer} show
photorealistic results, but lack explicit control over 3D properties~\cite{he2025unirelight} and
suffer from inconsistencies across views and camera trajectories due to their generative nature. Furthermore, these approaches remain computationally demanding and are often designed for offline use.

Our method combines the strength of explicit Gaussian representations and a neural renderer. To alleviate the ill-posed nature of intrinsic decomposition, we leverage 
existing priors \cite{DiffusionRenderer} and extend 3D Gaussian ray tracing~\cite{3dgrt2024} to dynamic PBR shading under novel HDR environment map lighting. Our pipeline is the first to render \emph{irradiance under dynamic geometry and illumination in real-time} for a Gaussian-based representation. This allows us to design a neural renderer conditioned on both PBR shading and irradiance, enabling dynamic control and improved realism over relightable Gaussian baselines. To mitigate the lack of paired real-world data, we propose a multi-stage training curriculum, including (i) large-scale pretraining on synthetic data with faithful illumination and material controls, (ii) domain adaptation to real-world scenes with imperfections, and (iii) multi-frame temporal consistency fine-tuning. To support interactive use, we adopt efficient single-step architectures~\cite{wu2025difix3d,zhang2026diffusionharmonizer}.

To evaluate our framework, we consider the task of novel-view rendering under modified relighting conditions in both synthetic and real-world settings. Our quantitative evaluation include visual error metrics to assess reconstruction quality and runtime costs, demonstrating clear improvements in both aspects. Beyond relighting, we showcase challenging real-world applications, including rendering \emph{dynamic shadows in a physically simulated Gaussian scene}, material editing and harmonized object insertion.
In summary, our contributions are as follows:
\begin{itemize}
    \item{A novel rendering framework that combines Gaussian ray tracing and a neural renderer for controllable, multi-view consistent, interactive rendering under dynamic conditions in real-world scenes}.
    \item{A multi-stage neural renderer training strategy that balances controllability and realism by combining large-scale synthetic pretraining, adaptation to real-world data with imperfect reconstructions, and temporal consistency optimization.}
    \item{Improved fidelity on real-world scenes in a range of interactive applications, including relighting, material editing, object insertion and removal.}
\end{itemize}

%% file: Sections/02_related.tex
\section{Related Work}
\label{sec:related}

Recent 3D reconstruction techniques synthesize near-photorealistic novel views \cite{mildenhall2020nerf, mueller2022instant, kerbl3Dgaussians}, and the $Lagrangian$ nature of particle-based primitives, such as Gaussians \cite{kerbl3Dgaussians, mai2024everexactvolumetricellipsoid, Huang2DGS2024, wu20253dgut, 3dgrt2024} and triangles \cite{govindarajan2025radfoam, Held2025Triangle+}, makes them especially appealing for simulating digital worlds. However, these reconstructed representations entangle materials and illumination into baked radiance, prohibiting editing and relighting,
even in the case of recently introduced Gaussian ray tracing techniques \cite{gao2024relightable,3dgrt2024,wu20253dgut}. Lighting and material decompositing techniques
targeting implicit representations
\cite{boss2021nerd, verbin2022refnerf, zhang2021NeRFactor, Mai2023NeuralMF, Jin2023TensoIR, zhang2023nemf, yao2022neilf} and signed distance functions \cite{physg2021, lin2025iris, liang2023envidr, zhang2023neilf++, zhang2022invrender} remain difficult to apply to
interactive editing.  While mesh extraction methods offer a solution, they require repeated conversions \cite{Munkberg_2022_CVPR, hasselgren2022nvdiffrecmc, wang2023fegr}, risk losing delicate details, or retain baked lighting \cite{guedon2023sugar, Yu2024GOF, Reiser2024SIGGRAPH, guedon2025milo}. Thus, we focus on extending dynamic lighting to Gaussian-based representations.

\paragraph{Relightable Gaussians.} A growing body of work addresses this limitation by optimizing Gaussian scene representations with partial physically based rendering (PBR) properties~\cite{R3DG2023,liang2023gs,wu2025gsssr,jiang2023gaussianshader,chen2025gigs,shi2025gir, liang2025gusir, ye2024progressiveradiancedistillationinverse}, emphasizing materials \cite{chung2025differentiable, zhang2025materialrefgs} or inter-reflections \cite{ye2024gsdr, poirierginter:hal-05306634, gu2024IRGS, han2026radiogs}. However, because inverse rendering is highly ill-posed, these works rely on fragile regularizations and implicit priors. This often causes shadow bleeding into the base color and severe degradation when relighting far from the original conditions. Consequently, many approaches resort to simplified domains like one-light-at-a-time (OLAT) setups \cite{bi2024rgs, fan2025rng, Dihlmann2024SSSGS, he2024metags}, single specular objects \cite{jiang2023gaussianshader, zhu_2025_gsror, zhou2025rtrgs, glossygs, gu2024IRGS, Wu2024DeferredGSDA, zhu2025discretizedsdf, ye2025geosplatting, sun2025svgir}, or specific outdoor scenes \cite{Haiyang2025GaRe, kaleta2025lumigauss, SuRGS_Zhang_2025_ICCV}.
While recent methods successfully leverage material priors from generative models \cite{zeng2024rgb, eftekhar2021omnidata, ke2023repurposing, noras2025gainsgaussianbasedinverserendering} for decomposition \cite{du2024gsidilluminationdecompositiongaussian, chen2025gigs}, dynamic illumination remains a bottleneck: by employing baked shadows or uniformly sampled shadow rays, these techniques fail to model long silhouettes and require reoptimization for dynamic geometry. Jiang et al.~\cite{jiang2025radiositygs} and Zhou et al.~\cite{zhou2024unified} set an elegant stage for differentiable global illumination, but are efficiency-limited to single objects or slow offline optimization, respectively. Orthogonal to our work, Poirier-Ginter et al.~\cite{poirierginter:hal-05306634} model indirect reflections but omit shadows and relighting. 
A universal limitation of this line of work is the reliance on constrained shading models, inherently precluding true photorealism. In contrast, our approach narrows this domain gap using a neural renderer while natively supporting dynamic shadows, respecting the material composition of the scene.

\textbf{Image-Space Neural Rendering.} A parallel line of work tries to achieve relighting through learning, for example by applying large generative models in image space \cite{lightlabMagar2025, erel2025practilight, kocsis2024iid, chen2024intrinsicanything, fortier2026spotlight, Xing2024luminet, zeng2024dilightnet, jin2024neural_gaffer, ICLR2025_iclight}, with some methods leveraging multi-view images~\cite{zhang2025relitlrm, PGZED19, litman2025lightswitch}, or videos~\cite{he2025unirelight, Xing2026GR3ENGR}. By training paired forward and inverse renderers, RGB$\leftrightarrow$X approaches~\cite{DiffusionRenderer, zeng2024rgb, lyu2025intrinsic} map rendered images to G-buffers and vice versa, achieving impressive intrinsic decomposition and relighting. While the generative nature of these approaches enables them to hallucinate realism, it limits their controllability and comes at a high computational cost. DIFIX3D~\cite{wu2025difix3d} and DiffusionHarmonizer~\cite{zhang2026diffusionharmonizer} present efficient pipelines for single-step inference by distilling pretrained diffusion models. 
Inspired by these approaches, we inject explicit 3D knowledge as conditions for neural rendering.

\paragraph{Hybrid Approaches.} %
Some multi-view reconstruction approaches independently relight each input image and lift to 3D \cite{alzayer2025generativemvr, bolduc2025GaSLight, PoirierGinter2024DiffusionRadianceRelighting}, or predict per-view materials \cite{litman2025materialfusion} and optimize a mesh reconstruction \cite{hasselgren2022nvdiffrecmc}. While promising, these works are limited in scene scale and suffer from hallucinated shadows and artifacts from inconsistent view predictions. Concurrent to our work, GR3EN~\cite{Xing2026GR3ENGR} uses a neural renderer to bake a new relit appearance into the original 3D reconstruction, precluding use under dynamic lighting or scene motion. Another recent work \cite{careagaRelighting} relights photographs by combining explicit path-tracing with a finetuned neural renderer in a 2.5D setting. To our knowledge, application of neural rendering for dynamic relighting of Gaussians has not been explored.

%% file: Sections/03_method_tracer.tex
\section{Method}\label{sec:overview}

\input{Figures/02_overview_architecture}

Given multi-view images and camera parameters $\left\{(I_i, C_i)\right\}_{i=1}^N$,
our goal is to infer a 3D representation that can be rendered under any novel global environment map $E$ or object motion.
Our framework (Fig.\ref{fig:overview}) optimizes a Gaussian particle representation (\S\ref{sec:gs_rep}) guided by existing priors (\S\ref{sec:prior_informed_rec}) and extends Gaussian ray tracing to PBR and irradiance (\S\ref{sec:tracing}), improving realism with a neural renderer that synthesizes the final image (\S\ref{sec:neural_renderer}). Our formulation supports dynamic rendering (no precomputation) under dynamic lighting, Gaussian object motion, insertion and removal, and PBR material editing.

\subsection{Gaussian Scene Representation}\label{sec:gs_rep}
We optimize a 3D scene $\mathcal{G}$ consisting of $M$ material-augmented Gaussians, defined as follows:

\begin{equation}
    \mathcal{G} = \left\{A_j = (\bmu_j, \bR_j, \bs_j, \sigma_j,\bbeta_j); \quad B_j = (\bb_j,m_j,r_j) \right\}_{j=1}^M,
\label{eq:gaussian-representation}
\end{equation}
where $A_j$ denotes standard 3DGS parameters \cite{kerbl3Dgaussians,3dgrt2024} (position $\bmu_j$, opacity $\sigma_j$, view-dependent color $\bbeta_j$ represented with spherical harmonics of degree $3$) and $B_j$ encodes our additional intrinsic material properties. To encourage better surfaces, we follow 2D Gaussian splatting\cite{Huang2DGS2024} and parametrize each Gaussian with a rotation matrix $\bR_j \in SO(3)$ and 2D scaling $\bs_j \in \mathbb{R}_{+}^2$. 
$B_j$ contains per-particle base color $\bb_j\in \mathbb{R}^3$, as well as scalar metallicity $m_j$, and roughness $r_j$ values. We leverage rich priors to disentangle these properties from the traditionally entangled appearance in $A_j$.

\subsection{Prior-Informed Reconstruction}\label{sec:prior_informed_rec}

To mitigate the ill-posed nature of material estimation from renderings alone, we leverage a learned intrinsic image decomposition prior \cite{DiffusionRenderer} to predict per-view material maps $\phi(I) = (I_n, I_b, I_r, I_m)$, containing normals $I_n$, base color $I_b$, roughness $I_r$ and metallicity $I_m$. These are collectively referred to as the G-Buffers, and define the augmented multi-view input $\left\{(I_i,\phi(I_i),C_i)\right\}_{i=1}^N$.
These predicted values provide a valuable signal, but can be inconsistent across views, making optimization more challenging. To stabilize optimization, we propose a two-stage procedure. 
In the first stage, we optimize the geometric and baked appearance $\left\{A_j\right\}$, supervised by images $I$ with an additional loss term between geometric normal and predicted $I_n$. This stage enforces alignment between geometry and appearance. In the second stage, we fix $\left\{A_j\right\}$ and optimize the materials $\left\{B_j\right\}$ using the remaining signals $(I_b, I_r, I_m)$. See \ref{sec:prior_informed_details} for full optimization details.

\subsection{Ray-Traced Shading and Irradiance}\label{sec:tracing}
Even with strong learned priors, reconstructed geometry and materials remain imperfect. As a result, directly using physically based rendering as the final output often produces artifacts. However, the reconstructed Gaussian scene still provides valuable 3D information about visibility, surface orientation, material composition, and light transport. 

Given the material-augmented Gaussian scene $\mathcal{G}$ and an environment map $E$, TRON extends Gaussian Ray Tracing~\cite{3dgrt2024,wu20253dgut} to render two guidance buffers: an approximate PBR image $\hat{I}_{\mathrm{pbr}}$ and a dynamic irradiance image $\hat{I}_{\mathrm{irr}}$. 
The PBR buffer captures local material response under the target illumination, while the irradiance buffer captures incoming light modulated by scene visibility, including shadows caused by dynamic geometry. These signals provide complementary information: $\hat{I}_{\mathrm{pbr}}$ encodes material-dependent appearance, whereas $\hat{I}_{\mathrm{irr}}$ exposes illumination and occlusion cues that are difficult for an image-space model to infer reliably.

Efficiently computing these buffers is challenging. Unlike mesh rendering, a primary ray may intersect many Gaussian primitives, making direct shading prohibitively expensive. We first address this with a deferred rendering strategy~\cite{deering1988deferredshading}: Gaussian contributions are first accumulated along each primary ray to form a representative surface point and G-buffer, and shading is then evaluated once per pixel rather than once per primitive.
For efficient PBR shading, we use the split-sum approximation~\cite{karis2013real,Munkberg_2022_CVPR}, which preintegrates environment lighting for material-dependent reflectance. Since this approximation does not fully account for visibility effects such as cast shadows, we additionally estimate irradiance by tracing a small number of secondary rays. We sample secondary directions using multiple importance sampling~\cite{veach1998montecarlo}, biasing samples toward high-contribution directions determined by both the material properties and the environment map. Finally, we exploit the order-invariance of transmittance to trace secondary rays out of order, enabling efficient execution on modern ray-tracing hardware. Additional implementation details are provided in Appendix~\ref{sec:gs_details}.

%% file: Figures/02_overview_architecture.tex
\begin{figure*}[t]
    \centering
    \includegraphics[width=0.95\textwidth]{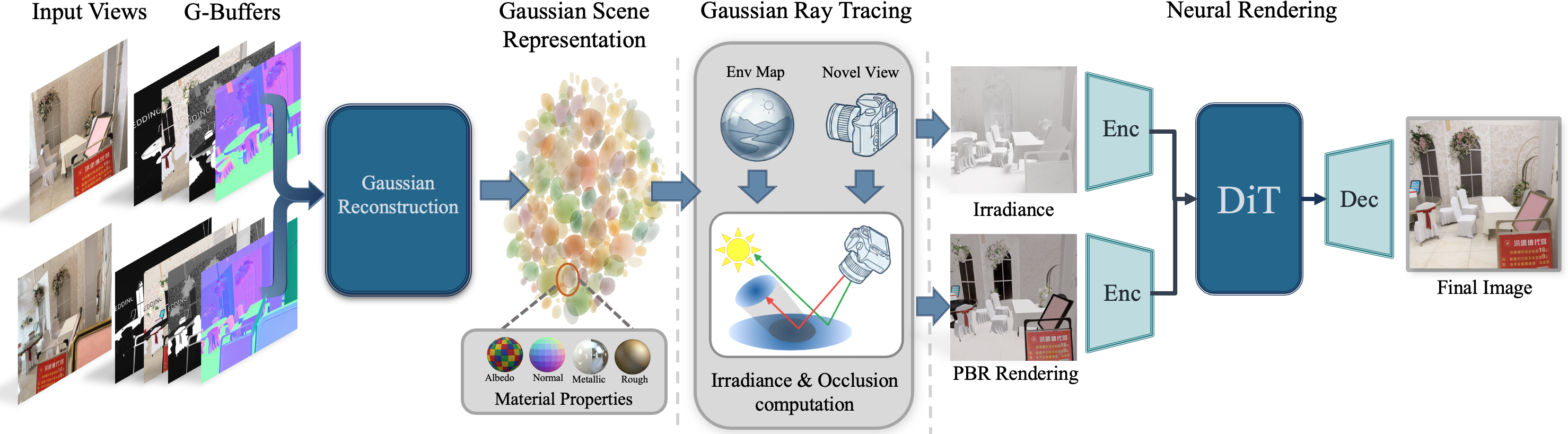}
    \caption{
        \textbf{TRON Architecture} 
        During training, we (1) apply an intrinsic decomposition model on all input view to get G-buffers and then (2) reconstruct a multiview scene as a set of 2D Gaussians, and (3) extract G-buffers diffusion priors per view, which we lift and bake in 3D. At inference time, given an envmap and camera view we (1) render a pair of buffers: PBR shaded and and irradiance. (2) Then the Neural Renderer maps these pairs of buffers to a high quality RGB output image. 
    }
    \label{fig:overview}
\end{figure*}

%% file: Sections/04_method_neural_renderer.tex
\section{Neural Renderer}
\label{sec:neural_renderer}

Our Gaussian ray tracer produces two image buffers per view: a physically based rendering, $\hat{I}_{\mathrm{pbr}}$, that captures direct light--material interactions, and an irradiance buffer, $\hat{I}_{\mathrm{irr}}$, that captures the incoming light at the surface including shadows. By construction, these buffers are approximations of a real photograph: (i)~they omit higher--order rendering effects, e.g., subsurface scattering, transparent and complex materials and (ii)~they inherit the limitations of Gaussian reconstruction, which typically covers only the foreground of a scene at high density and leaves the periphery sparse, blurry, or missing. For those reasons, closing this gap with a classical forward renderer is impractical. Instead, we propose a \emph{neural renderer} $\mathcal{R}_\theta$ conditioned on $\hat{I}_{\mathrm{pbr}}$ and $\hat{I}_{\mathrm{irr}}$ which produces a single realistic RGB image $\hat{I}_{\mathrm{rgb}}\!=\!\mathcal{R}_\theta(\hat{I}_{\mathrm{pbr}},\hat{I}_{\mathrm{irr}})$. We treat $\mathcal{R}_\theta$ as a strongly conditioned generative model: it must (a)~be expressive enough to hallucinate the missing rendering effects and inpaint reconstruction artifacts, and (b)~be efficient enough to be queried at interactive rates in a streaming setting. We first describe the architecture and training objective, and then our paired data curation strategy.

\subsection{Architecture}\label{sec:neural_renderer:arch}

\textbf{Backbone.} Pretrained text-to-image diffusion models contain strong generative priors that have been shown to transfer well to image-to-image translation tasks~\citep{zhang2026diffusionharmonizer}. We adopt the same recipe and build on Cosmos 0.6B~\citep{nvidia2025cosmosworldfoundationmodel}, a text-to-image latent diffusion model that combines a $0.6$ billion parameter DiT~\citep{nvidia2025cosmosworldfoundationmodel} with a $0.14\mathrm{B}$-parameter WAN-2.1 causal video autoencoder~\citep{wan2025} as its tokenizer. The autoencoder $(\mathcal{E},\mathcal{D})$ consumes either an image or a short video and emits a temporally compressed latent that the DiT $\mathcal{F}_\theta$ denoises in latent space. To adapt the model to our neural rendering task, we finetune the DiT from the public Cosmos weights; the encoder and decoder are kept frozen. %

\textbf{Input encoding.} Because the two input buffers carry very different statistics $\hat{I}_\mathrm{pbr}$ is an PBR rendering that includes high-frequency details, textures and reflections, 
while $I_{\mathrm{irr}}$ is a smooth low-frequency irradiance map, 
we encode the two buffers separately and average the result to create our input latent:
\begin{equation}
z = \tfrac{1}{2}\bigl(\mathcal{E}(\hat{I}_{\mathrm{pbr}}) + \mathcal{E}(\hat{I}_{\mathrm{irr}})\bigr).
\label{eq:fusion}
\end{equation}
Compared to channel-wise concatenation, averaging keeps the latent dimensionality of the original Cosmos backbone unchanged, allowing us to reuse all DiT weights. Figures~\ref{fig:ablation_latent_weight_grid_2} and~\ref{fig:ablation_latent_weight_grid_1} present an ablation study of this design choice. 

\textbf{Single-step deterministic denoising.} In standard latent diffusion, the backbone $\mathcal{F}_\theta$ is queried for many timesteps along a stochastic noise schedule and conditioned on a text prompt. For real-time online rendering this is prohibitively expensive. Following the single-step training paradigm of~\citep{zhang2026diffusionharmonizer}, we repurpose $\mathcal{F}_\theta$ into a deterministic image-to-image operator: we feed the clean latent $z$ from Eq.~\eqref{eq:fusion} directly into the DiT, fix the diffusion timestep to a constant value $\tau$ (we use $\tau=250$ on a $1000$-step schedule), and replace the text-conditioning tokens with the model's null embedding. The decoder then produces the predicted RGB image:
\begin{equation}
I_{\mathrm{rgb}} \;=\; \mathcal{D}\!\Bigl( \mathcal{F}_\theta\!\bigl(z;\, \tau,\, \varnothing\bigr) \Bigr).
\label{eq:renderer}
\end{equation}
This single forward pass yields a stable, deterministic mapping from the input buffers to RGB, eliminates iterative sampling, and improves frame-to-frame structural consistency in the streaming setting.

\textbf{Temporal context.} A purely per-frame renderer can flicker on moving cameras because the network is free to choose a different realization of the missing content each frame. We therefore extends $\mathcal{R}_\theta$ to operate on short clips. Let $K$ be the clip length; following the WAN-2.1 tokenizer, which compresses time by a factor of $4$, we use $K=1{+}4n$ ($K{=}5$ in practice). We stack the $K$ frames of each buffer along the temporal axis and run a single spatio-temporal forward pass:
\begin{equation}
z_{t:t+K-1} \;=\; \tfrac{1}{2}\!\bigl( \mathcal{E}(I_{\mathrm{pbr},\,t:t+K-1}) + \mathcal{E}(I_{\mathrm{irr},\,t:t+K-1})\bigr),
\label{eq:fusion_temporal}
\end{equation}
which is then mapped by the DiT and decoded as in Eq.~\eqref{eq:renderer}. The DiT activates its native interleaved spatial and temporal attention layers, so it can use neighbouring frames as context for ambiguous regions while keeping the per-frame structure of $\hat{I}_{\mathrm{pbr}}$ intact. For longer trajectories we apply $\mathcal{R}_\theta$ in a sliding window of $K$ frames with an overlap of $K-1$ (a stride of one) and resolve overlapping predictions by averaging or by discarding non-anchor frames.

\textbf{Training objective.} We supervise $\mathcal{R}_\theta$ against the ground-truth RGB frames $I^{\star}$ with a pixel and a perceptual term, applied independently to every frame of the clip:
\begin{equation}
\mathcal{L} \;=\; \lambda_{2}\,\bigl\|\hat{I}_{\mathrm{rgb}} - I^{\star}\bigr\|_2^2 \;+\; \lambda_{p}\,\mathcal{L}_{\mathrm{LPIPS}}\!\bigl(\hat{I}_{\mathrm{rgb}},\,I^{\star}\bigr).
\label{eq:loss}
\end{equation}
The $\ell_2$ term anchors the predicted radiance to the correct mean and matches global tone, exposure, and colour balance, but on its own tends to produce blurry predictions wherever the input-to-RGB mapping is multi-modal (e.g.\ specular highlights or missing regions). The LPIPS term~\cite{lpips}, a weighted distance in the feature space of a frozen VGG-16~\cite{Simonyan15} model, counters this by allowing the network to commit to plausible high-frequency content instead of averaging over it, and is the main driver of sharp textures and crisp edges. We use $\lambda_2{=}\lambda_p{=}1.0$ throughout, and the same loss is applied in both the image and video stages; only the DiT $\mathcal{F}_\theta$ receives gradient updates.

\subsection{Paired Data Curation}
\label{sec:neural_renderer:training}

Training $\mathcal{R}_\theta$ requires paired samples $\{(\hat{I}_{\mathrm{pbr}},\,\hat{I}_{\mathrm{irr}},\,I^{\star})\}$ in which the two input buffers come from the Gaussian ray tracer and the target $I^{\star}$ is a high-quality photoreal frame of the same view. Such triples are not available off-the-shelf, so we build a curated mixture of synthetic and real data.

\textbf{Synthetic data.} For full ground-truth control we render $975$ procedurally varied scenes through a path tracer, each from a fixed $93$-frame camera trajectory and under $20$ distinct illumination settings ($10$ high-contrast and $10$ low-contrast environment maps), and pair every rendered frame with the matching PBR and irradiance buffers produced by the Gaussian Ray Tracer on the same scene. This yields $1.81\,\mathrm{M}$ paired samples spanning a wide range of materials, geometries, and light transports.

\textbf{Real data.} To close the synthetic--real gap we additionally curate $955$ real captures from the DL3DV~\citep{ling2024dl3dv} benchmark, reconstructed with our Gaussian representation and optimized for material (Section~\ref{sec:prior_informed_rec}) and envmap (Appendix \ref{sec:app:envmap_optimization}). Each scene contributes on average $\sim\!300$ frames at its native illumination, for a total of around $ 290\,\mathrm{K}$ paired samples. Real captures introduce reconstruction artifacts (sparsity at the periphery, view-dependent specularities, dynamic exposure) that are absent from the synthetic split and are essential for transferring the renderer to in-the-wild content.

\textbf{Three-stage curriculum.} We train the DiT $\mathcal{F}_\theta$ in three stages, with the encoders and decoder held frozen throughout. \emph{Stage~1:} we use single-frame samples ($K{=}1$) drawn exclusively from the synthetic split and optimize the loss in Eq.~\eqref{eq:loss}. This stage exposes $\mathcal{F}_\theta$ to a wide range of materials and illuminations under clean ground truth and adapts the diffusion prior to the buffer statistics. \emph{Stage~2:} we keep $K{=}1$ but switch the data source to the real split. This stage narrows the domain gap and teaches $\mathcal{F}_\theta$ to handle the reconstruction artifacts characteristic of in-the-wild captures. \emph{Stage~3:} we move to $K{=}5$ fixed-length clips on the real split, which exercises the DiT's temporal attention layers under the same per-frame supervision. This stage teaches $\mathcal{F}_\theta$ to leverage temporal context for flicker-free streaming without drifting from the per-frame solution learned in Stages~1--2. The combined training set comprises approximately $2.1\,\mathrm{M}$ paired samples.

%% file: Sections/05_evaluation.tex
\input{Figures/combined_materials}

\input{Figures/fig_compare_relight_ALL}
\input{Tables/combined_quant_relight}

\section{Experiments}
\label{sec:evaluation}

We evaluate \ourmethod{} on a mix of synthetic and real-world datasets. 
In our quantitative evaluation (\S\ref{sec:eval:quant}), we focus on fidelity of material decomposition, realism of relighting, and rendering speed. Our qualitative results (\S\ref{sec:eval:qual}) support these findings and highlight the novel capabilities of our method, difficult to achieve with other techniques such as harmonization, precise control of shadows under dynamic scene motion, material editing and high quality interactive preview.
Throughout, we emphasise that our objective is to plausibly relight a scene under \emph{novel} illumination rather than to faithfully reproduce its original captured appearance, a setting at which existing reconstruction methods already excel.

\subsection{Experimental Setup}

\paragraph{Benchmark.} To evaluate our material decomposition and relighting, we extend the evaluation benchmark of previous works~\cite{DiffusionRenderer} and generate a synthetic test dataset, we name \textit{\tronsynth{}}, using the same path tracing pipeline as in our synthetic training data generation (See Appendix~\ref{sec:dataset_details}) but with a different set of 3D objects, environment maps and render settings. Our test dataset contains 15 object-centric scenes and 25 multi-object scenes, each rendered with a path tracer under 5 sampled illuminations, and paired with ground-truth material decompositions.  For completeness with prior art, we also evaluate on the TensorIR~\cite{Jin2023TensoIR} dataset, consisting of 4 shapes with their materials, each relit with 5 different envmaps. %
For evaluation on real data we denote \textit{\dldvbenchmark{}} as a dataset consisting of 33 scenes from the 2nd partition of DL3DV-10K \cite{ling2024dl3dv}. These scenes were never used during training, and each one is paired with 3 different rotated environment maps from a curated collection of envmaps consisting of different contrast levels.
\vspace{-5pt}
\paragraph{Baselines.} We evaluate \ourmethod{} against both Gaussian-based relighting methods and neural image/video models. Note that we exclude relightable NeRF variants, which are not well-suited for dynamic scene motion or local edits, our target application area.
We select GaussianShader \cite{jiang2023gaussianshader}, GS-IR \cite{liang2023gs} and GI-GS \cite{chen2025gigs} to represent relightable reconstruction methods, given their ability to model occlusions, extend to scene scale, and public code available. 
We also include published results from GS-ID\cite{du2024gsidilluminationdecompositiongaussian}, which to the best of our knowledge, is the only Gaussian based method using material priors for relighting, but has no public codebase. Furthermore, we compare against DiffusionRenderer~\cite{DiffusionRenderer} and UniRelight~\cite{he2025unirelight}, which represent neural rendering and relighting, respectively.
We examine how both categories of methods behave under novel views with varying illuminations.

\vspace{-5pt}
\subsection{Quantitative Evaluation and Comparisons} \label{sec:eval:quant}\label{sec:eval:compare}

\paragraph{Material Decomposition.} We use synthetic datasets to evaluate the fidelity of our material decomposition in Tb.\ref{fig:qual:mat}, reporting PSNR, SSIM and LPIPS \cite{lpips} for rendered albedo (scaled according to common practice, see Appendix \ref{app:albedo_scaling}) and Mean Angular Error (MAE) for rendered normals. By utilizing robust priors \cite{DiffusionRenderer},  
\ourmethod{} strongly outperforms Gaussian relighting baselines that optimize for material decomposition by matching original scene appearance, often resulting in artifacts like shadows baked into albedo (Fig.\ref{fig:qual:mat}) and normals that struggle to converge in regions with intricate geometry or poor camera coverage. On the other hand, by baking priors into an existing high-quality 3D scaffold, our method avoids resolution limitations and consolidates any multi-view inconsistencies exhibited by neural approaches, explaining our improved performance over this class of approaches along several metrics. In contrast to neural techniques, our baked materials are 3D consistent and can be rendered in real-time from any view under dynamic scene changes.  Though we share similarities with previous work by GS-ID \cite{du2024gsidilluminationdecompositiongaussian} and use diffusion priors to supervise geometry, our 2D Gaussian representation is also supervised through SSIM loss (Eq.~\ref{eq:tron_recon_loss}), allowing prior knowledge to backpropagate through the geometric normals directly into the Gaussian scale and rotation parameters, where SSIM loss helps preserve the high-frequency details present within the prior.

\input{Figures/fig_tron_gallery}
\input{Figures/fig_harmonize_compare}

\vspace{-8pt}
\paragraph{Relighting.} 
We compare our method against baselines on the task of \emph{novel view synthesis under relighting}. No ground truth dataset for this task exists for complex real-world scenes, our target domain. Therefore we use approximate metrics to assess performance on real scenes in \dldvbenchmark{}. To compare \textbf{photorealism}, we
devise an explainable agent-based metric that uses a VLM to perform pairwise comparisons between frames relit by our method and the baselines ; see Appendix~\ref{app:agent_metrics} for prompt and examples. For each scene and relighting condition, we sample 5 views, evaluating the win rate of our method against every method across all 495 pairs per method with 3 different seeds for standard deviation. Our results are summarized in Tb.~\ref{tab:photorealism}.
As a sanity check, we also compare our method against the original frames, and validate real frames win 89\% of the time. Our method significantly outperforms Gaussian-based baselines, achieving win rate of 97\%-99\%. While these methods can achieve competitive PSNR on original scene reconstruction with optimized materials, their entangled materials (Fig.\ref{fig:qual:mat})
result in degraded appearance under novel lighting (Fig.\ref{fig:relight_compare}) of in-the-wild captured scenes. For neural inverse rendering techniques, trained to hallucinate highly
realistic output, we observe mixed results with win rates of 47\%-64\%. This performance of more computationally heavy neural techniques comes at a cost of much lower controllability (for example, with strong shadows mismatches for different rendered trajectories of the same scene. See Fig.\ref{fig:randomseed},
and offline rendering speed. The last column in Tb.\ref{tab:photorealism} presents elapsed time for feedback, e.g. latency from the moment rendering begins to the first frame appears. Our entire pipeline, tracer and neural renderer included, has a latency of under a second, which allows our method to be directly used for user-driven interactive applications, where immediate response to edits is required. 

Furthermore, we validate our method on the Video Generative Models benchmark of VBench \cite{zhang2024evaluationagent}, with 5 criteria that allow evaluation directly on rendered videos without an accompanying prompt. Summarized in Tb.\ref{tab:vbench}, we compare smooth
trajectories, using the first 100 frames per relit variation of a scene in \dldvbenchmark{}, resized to $1920 \times 1080$) rendered by our method and neural baselines, totalling in comparison over 99 unique clips. Our method shows competitive performance with these offline, full denoising pipelines, while exhibiting interactive speed, controllability, and opening up new dynamic applications for captured 3D worlds. We also point out differences in how our method and \cite{DiffusionRenderer, he2025unirelight} handle shadows: the group of neural renderers relies on prior knowledge embedded in their weights, whereas ours is guided by irradiance signals grounded in a 3D world. Consequently, these methods have limited control over where shadows appear, possibly hallucinating unseen occluders (Fig.~\ref{fig:harmonization}) or silhouettes not faithful to subtle scene geometry (Fig.~\ref{fig:relight-syntheticB-neural}).

\vspace{-11pt}
\paragraph{More Results and Applications}\label{sec:eval:qual}

The combination of a 3D grounded representation (Section~\ref{sec:gs_rep}) and an informed neural renderer (Section~\ref{sec:neural_renderer}) unlocks a variety of applications, including harmonization, material editing (Fig.~\ref{fig:harmonization}), and rendering shadows of moving objects under changing lighting conditions (Fig.~\ref{fig:editing_gallery}), previously challenging to achieve with Gaussian based relighting methods. See the accompanying video for additional results.

%% file: Figures/combined_materials.tex
\begin{figure*}[ht]
\vspace{5pt}
\centering

\begin{minipage}{\linewidth}
\centering
\includegraphics[width=\linewidth]{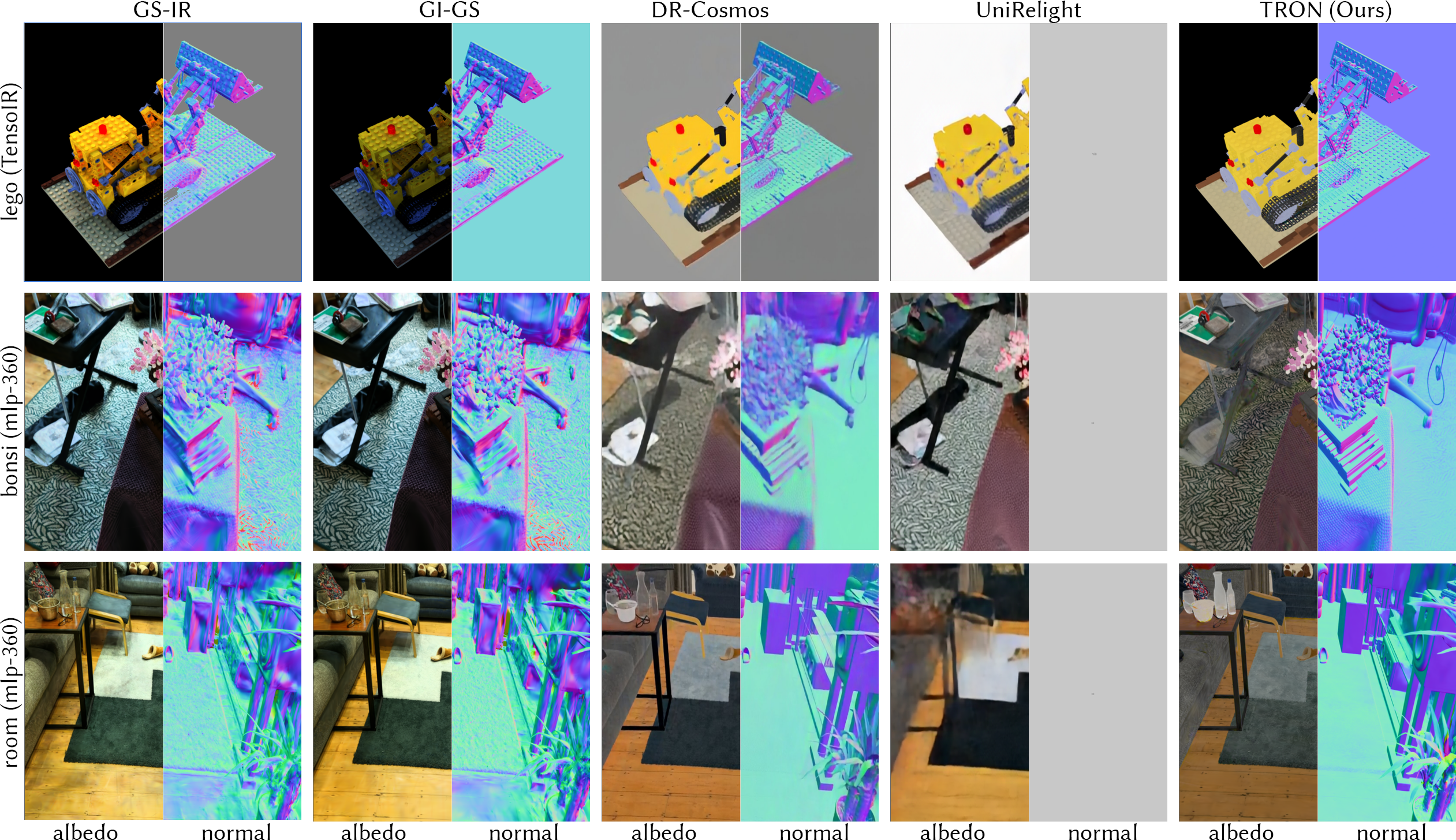}
\vspace{2pt}
\captionof{figure}{\textbf{Qualitative Material Decompositions Comparison}. As inverse-rendering is an ill-posed problem, previous Gaussian based methods \cite{chen2025gigs, liang2023gs} that rely on direct optimization typically suffer from shadows baked into the albedo layer. Neural rendering methods \cite{DiffusionRenderer, he2025unirelight} either sacrifice spatial resolution to maintain a temporal context window (in the case of video-based models), or suffer from multi-view inconsistencies. By baking priors as a volumetric feature field, TRON can consolidate inconsistencies from high resolution priors.}\label{fig:qual:mat}
\end{minipage}

\vspace{6pt}

\begin{minipage}{\linewidth}
\centering
\small
{
\begingroup
\renewcommand{\arraystretch}{1.3}
\setlength{\tabcolsep}{4pt}
\resizebox{\linewidth}{!}{
\begin{tabular}{l|cccc|cccc|cccc}
\Xhline{2.0pt}
 & \multicolumn{4}{c|}{TensorIR}
 & \multicolumn{4}{c|}{{\tronsynth} Objects}
 & \multicolumn{4}{c}{{\tronsynth} Scenes} \\ [1.5ex]
 \cline{2-13}
 \textbf{Method} & \multicolumn{3}{c}{\textbf{Albedo}} & \multicolumn{1}{c|}{\textbf{Normals}}
 & \multicolumn{3}{c}{\textbf{Albedo}} & \multicolumn{1}{c|}{\textbf{Normals}}
 & \multicolumn{3}{c}{\textbf{Albedo}} & \textbf{Normals} \\ [1ex]
 \cline{2-13}
 & PSNR$\uparrow$ & SSIM$\uparrow$ & LPIPS$\downarrow$ & MAE$\downarrow$
 & PSNR$\uparrow$ & SSIM$\uparrow$ & LPIPS$\downarrow$ & MAE$\downarrow$
 & PSNR$\uparrow$ & SSIM$\uparrow$ & LPIPS$\downarrow$ & MAE$\downarrow$ \\ [1ex]
\Xhline{2.0pt}
GS-IR \cite{liang2023gs}
 & 30.06 & 0.928 & 0.107 & 5.32
 & 19.85 & 0.831 & 0.305 & 10.85
 & 19.82 & 0.841 & 0.316 & 11.69 \\
GI-GS \cite{chen2025gigs}
 & 29.64 & 0.927 & 0.107 & \cellcolor{yellow!40}5.25
 & 19.10 & 0.797 & 0.316 & \cellcolor{yellow!40}10.17
 & 18.84 & 0.814 & 0.338 & \cellcolor{yellow!40}11.22 \\
GaussianShader \cite{jiang2023gaussianshader}
 & 23.55 & 0.832 & 0.168 & 14.90
 & 9.23 & 0.356 & 0.533 & 34.15
 & 8.54 & 0.376 & 0.540 & 37.37 \\
GS-ID (\textit{paper}) \cite{du2024gsidilluminationdecompositiongaussian}
 & \cellcolor{orange!35}33.49 & \cellcolor{yellow!40}0.952 & \cellcolor{yellow!40}0.079 & \cellcolor{orange!35}4.602
 & -- & -- & -- & --
 & -- & -- & -- & -- \\
\Xhline{1.0pt}
DR-Cosmos \cite{DiffusionRenderer} \textit{(2D prior)}
 & \cellcolor{yellow!40}33.22 & \cellcolor{orange!35}0.959 & \cellcolor{orange!35}0.073 & 5.34
 & \cellcolor{red!30}\textbf{30.99} & \cellcolor{red!30}\textbf{0.940} & \cellcolor{red!30}\textbf{0.156} & \cellcolor{orange!35}7.19
 & \cellcolor{red!30}\textbf{31.11} & \cellcolor{red!30}\textbf{0.940} & \cellcolor{red!30}\textbf{0.131} & \cellcolor{orange!35}5.59 \\
UniRelight \cite{he2025unirelight} 
 & 32.69 & 0.947 & 0.089 & --
 & \cellcolor{yellow!40}27.36 & \cellcolor{yellow!40}0.854 & \cellcolor{yellow!40}0.301 & --
 & \cellcolor{yellow!40}27.98 & \cellcolor{yellow!40}0.858 & \cellcolor{yellow!40}0.273 & -- \\
\Xhline{1.0pt}
\textbf{Ours}
 & \cellcolor{red!30}\textbf{34.10} & \cellcolor{red!30}\textbf{0.964} & \cellcolor{red!30}\textbf{0.069} & \cellcolor{red!30}\textbf{4.37}
 & \cellcolor{orange!35}29.65 & \cellcolor{orange!35}0.933 & \cellcolor{orange!35}0.250 & \cellcolor{red!30}4.97
 & \cellcolor{orange!35}31.09 & \cellcolor{orange!35}0.925 & \cellcolor{orange!35}0.261 & \cellcolor{red!30}3.58 \\
\Xhline{2.0pt}
\end{tabular}
}
\endgroup
}
\captionof{table}{\textbf{Quantitative Material Decompositions Comparison}. By lifting priors to 3D and baking them into the Gaussian field, \ourmethod{} consolidates high frequency multiview inconsistencies, improving in most cases over the priors (DR-Cosmos \cite{DiffusionRenderer}).}\label{fig:quan:mat}
\end{minipage}
\end{figure*}

%% file: Figures/fig_compare_relight_ALL.tex
\begin{figure*}[ht!]
\centering
\includegraphics[width=\linewidth]{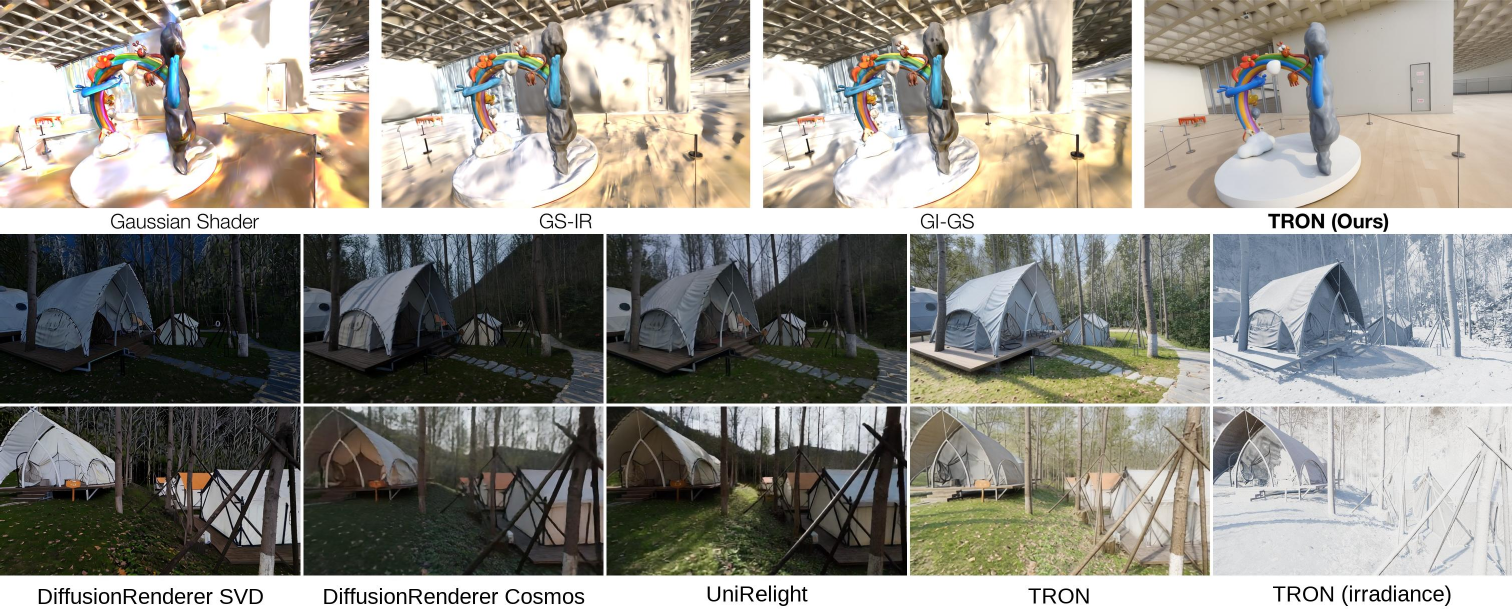}
\caption{
\textbf{Qualitative Relighting Comparisons}. 
Top: Gaussian Relighting baselines. To achieve interactive-speed shading, previous Gaussian-based relighting methods apply split-sum \cite{karis2013real}, which uses convolved lighting and is incompatible for modeling occlusions. TRON separates pbr and irradiance channels to avoid these pitfalls. Bottom: Neural Rendering baselines. Contemporary neural methods rely on priors embedded in their weights to hallucinate realistic shadows, while \ourmethod{} respects explicit 3D structure (last column).} \label{fig:relight_compare}
\end{figure*}

%% file: Tables/combined_quant_relight.tex
\begin{figure*}[ht]
\vspace{-11pt}
\centering
\begin{minipage}{0.53\linewidth}
\centering
\small
{
\begingroup
\resizebox{\linewidth}{!}{
\begin{tabular}{l c c c}
    \toprule
    V.S. Baseline                   & Ours Win Rate & FPS $\uparrow$ & First-frame \\
    & & & Latency $\downarrow$ \\
    \midrule
    GaussShader             & 99.73 $\pm$ 0.12 & 21.06 & 47.5\,ms \\
    GI-GS                      & 97.71 $\pm$ 0.42 & 3.64  & 275\,ms \\
    GS-IR                      & 97.31 $\pm$ 0.31 & 15.87 & 63\,ms \\
    \midrule
    DR-SVD      & 64.92 $\pm$ 0.62 & 0.54  & 44.7\,s \\
    DR-Cosmos   & 50.03 $\pm$ 0.65 & 0.96  & 59.7\,s \\
    UniRelight                 & 47.27 $\pm$ 1.41 & 0.13  & 450.0\,s \\
    \midrule
    TRON (PBR)            & 89.36 $\pm$ 1.49 & \textbf{51.81} & \textbf{19.3\,ms} \\
    TRON (PBR $\times$ Irrad.)  & 74.61 $\pm$ 0.84 & 8.90 & 112.3 ms \\
    TRON Full        & -- & 1.6 & 625\,ms  \\
    GT (real photo)            & 10.84 $\pm$ 0.62 & -- & -- \\
    \bottomrule
  \end{tabular}  
  }
\endgroup
}
\vspace{7pt}
\captionof{table}{\textbf{Relighting A/B Photorealism Comparison}: \ourmethod{} win-rate (\%) of our method vs.\ all baselines, as judged by a GPT 5.1 agent.
}\label{tab:photorealism}
\end{minipage}%
\hfill %
\begin{minipage}{0.45\linewidth}
\centering
\small
{
\begingroup
\renewcommand{\arraystretch}{1.1}
\setlength{\tabcolsep}{4pt}
\resizebox{\linewidth}{!}{
\begin{tabular}{l|ccccc}
\Xhline{2.0pt}
 & \multicolumn{5}{c}{\textbf{Metrics}} \\ [1.5ex]
 \cline{2-6}
 \textbf{Method}
 & \rotatebox{90}{Motion Smoothness} 
 & \rotatebox{90}{Subject Consistency} 
 & \rotatebox{90}{Background Consistency} 
 & \rotatebox{90}{Aesthetic Quality} 
 & \rotatebox{90}{Imaging Quality} \\ [1ex]
\Xhline{2.0pt}
DR-SVD      & 0.9820 & 0.8658 & 0.9088 & \cellcolor{orange!35}0.4891 & \cellcolor{orange!35}0.6263 \\
DR-Cosmos  & 0.9839 & 0.8617 & 0.9141 & 0.4707 & 0.5895 \\
UniRelight   & \cellcolor{red!30}\best{0.9865} & \cellcolor{red!30}\best{0.8738} & \cellcolor{orange!35}0.9160 & 0.4680 & 0.5522 \\
\Xhline{1.0pt}
\textbf{TRON (Ours)} & \cellcolor{orange!35}0.9787 & \cellcolor{orange!35}0.8737 & \cellcolor{red!30}\best{0.9190} & \cellcolor{red!30}\best{0.4959} & \cellcolor{red!30}\best{0.6956} \\
\Xhline{2.0pt}
\end{tabular}
}
\endgroup
}
\captionof{table}{\textbf{Quantitative Video-Quality Assessment:} measured via VBench \cite{zhang2024evaluationagent} on %
illuminated clips from \dldvbenchmark{}. Higher score is better.%
}\label{tab:vbench}
\end{minipage}

\end{figure*}

%% file: Figures/fig_tron_gallery.tex
\begin{figure*}[t!]
\centering
\includegraphics[width=\linewidth]{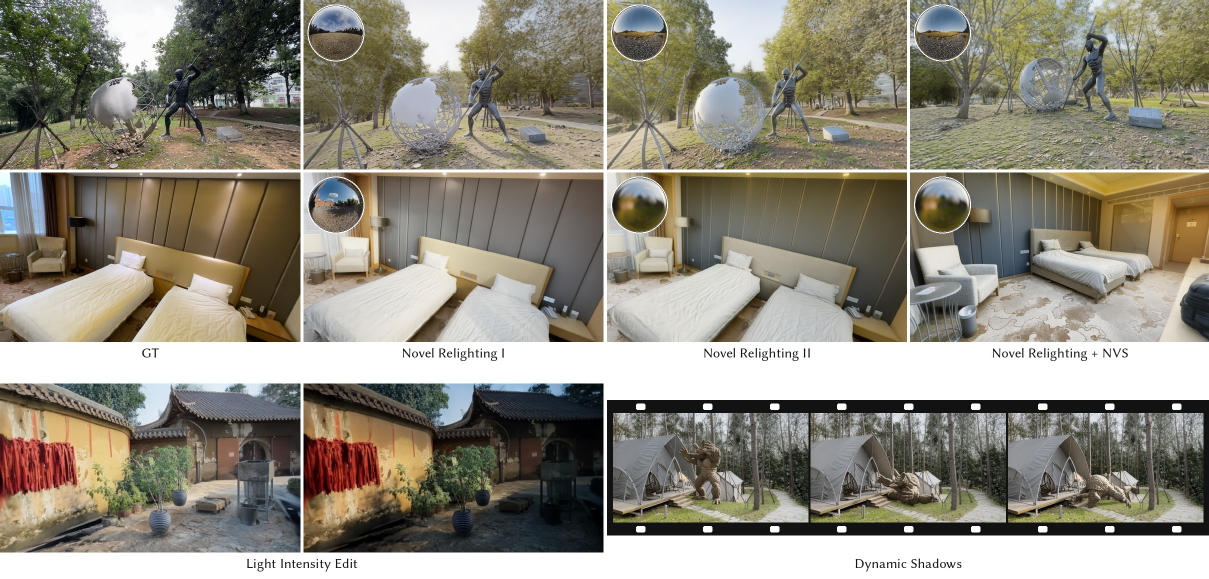}
\caption{\textbf{Results Gallery.} Top \& middle: show \ourmethod{} results for Novel Relighting on real-world captures. Bottom: editing global light intensity (left) and rendering shadows of a dynamic harmonized object (right).} \label{fig:editing_gallery}
\end{figure*}

%% file: Figures/fig_harmonize_compare.tex
\begin{figure*}[t!]
\vspace{-10pt}
\centering
\includegraphics[width=\linewidth]{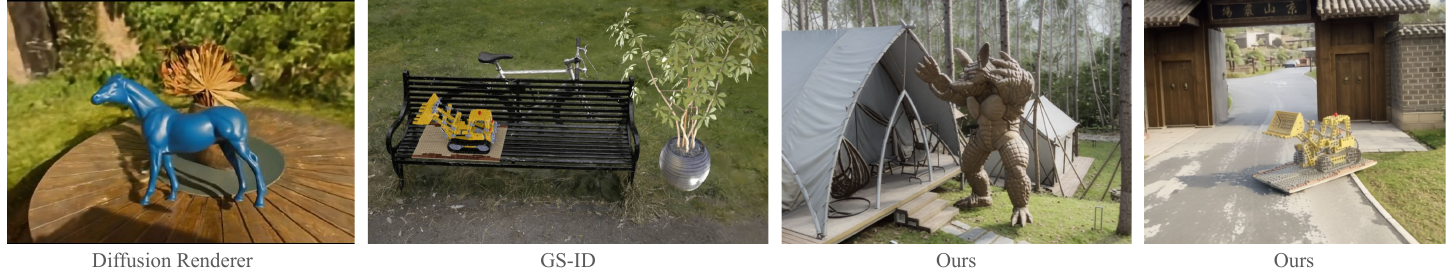}
\caption{\textbf{Harmonization Comparisons}: Our method dynamically computes the highlight and shadow conditions of the scene for a harmonized object.}\label{fig:harmonization}
\end{figure*}

%% file: Sections/07_limitations_conclusion.tex
\vspace{-5pt}
\section{Conclusion \& Limitations}
\label{sec:conclusion}
\vspace{-3pt}
We presented \ourmethod{}, a novel rendering pipeline combining Gaussian ray tracing with neural rendering to achieve a fast, controllable and high quality renderings of real-world captures.
While our approach enables novel interactive editing, its final image quality is bounded by the fidelity of the underlying 3D reconstruction, degrading when training view coverage is sparse and signals become noisy. Specifically, background regions are underdetermined and are prone to temporal artifacts. Moving forward, this interactive framework opens exciting avenues for future work, particularly in enforcing better object identity preservation within the neural renderer, further achieving real time performance and accurately modeling complex, 3D-aware phenomena like refractions and better control of near field illumination.

\section*{Acknowledgements}
We thank Nicolas Moenne-Loccoz, Yuxuan (Alex) Zhang, Miloš Hašan, Zheng Zeng, Jon Hasselgren, Vismay Modi, Sai Bangaru, Zan Gojcic, and Nicholas Sharp for valuable discussions and for their support with the prior work this paper builds upon.

%% file: Sections/appendix.tex
\section{Additional Results}

\input{Figures/fig_random_seed_comparison}
\vspace{20pt}
\input{Figures/appendix/dl3dv_material_decomposition}

\include{Figures/appendix/relight-syntheticB-gauss}
\include{Figures/appendix/relight-syntheticB-neural}
\input{Figures/appendix/mip360_gaussian_comparison}

\input{Figures/fig_ablation_latent_grid}

Fig.~\ref{fig:dl3dv_material_decomp} presents an additional comparison between Gaussian baselines, and \ourmethod{} with and without the neural renderer. Qualitative comparisons over \tronsynth{} are provided in Fig.~\ref{fig:relight-syntheticB-gauss} and Fig.~\ref{fig:relight-syntheticB-neural}. 
We ablate our Gaussian ray tracer v.s. the full pipeline on mip-360 in Fig~\ref{fig:mip360comp}.
In Fig.\ref{fig:ablation_latent_weight_grid_2} and Fig.\ref{fig:ablation_latent_weight_grid_1} we present a sweep ablation on weighting the neural renderer latents, showing how the different guidance signals affect the final outcome.

\section{Preliminaries}

\subsection{3D Gaussian Ray Tracing}
\label{sec:preliminaries}

\tdgrt{}~\cite{3dgrt2024} models scene geometry and appearance using a set of semi-transparent \textit{radiating} Gaussian particles. The spatial influence of each Gaussian at a point $\mathbf{x} \in \mathbf{R}^3$ is defined by the kernel function:
\begin{equation}
\label{eq:3dgrt_kernel}
  G_i(\mathbf{x})=\exp\left(-\frac{1}{2}(\mathbf{x}-\mu_i)^T\Sigma_i^{-1}(\mathbf{x}-\mu_i)\right),  
\end{equation}
where the mean position $\mathbf{\mu}_i$ and the anisotropic covariance matrix $\Sigma_i$ are optimizable, allowing the underlying Gaussian distribution to flexibly take the approximated form of a 2D surflet or 3D volume. To evaluate the radiance along a given ray $\mathbf{r}(t) = \mathbf{o} + t\mathbf{d}$, 3DGRT relies on an approximation of the Volumetric Rendering Equation:

\begin{equation}
\label{eq:3dgrt_volumetric}
L(\mathbf{o},\mathbf{d})\approx\sum_{i=1}^{N}\mathbf{c}_i(\mathbf{d})\alpha_i \prod_{j=1}^{i-1}(1-\alpha_j)
\end{equation}

Here, opacity $\alpha_i = \sigma_i G_i (\mathbf{x})$ is linearly approximated at $\mathbf{x}(t)$, the point along ray $\mathbf{r}$ maximizing the Gaussian response $G_i(\mathbf{x})$.

\subsection{Physically Based Rendering} 

The rendering equation \cite{10.1145/15886.15902} formulates the relationship between the outgoing radiance and incident light:

\begin{equation}
\label{eq:rendering_equation}
L_o(\mathbf{x}, \mathbf{\omega}_o) = \int_{\Omega} f_r(\mathbf{x}, \mathbf{\omega}_o, \mathbf{\omega}_i) L_i(\mathbf{x}, \mathbf{\omega}_i) |\mathbf{n} \cdot \mathbf{\omega}_i| \, d\mathbf{\omega}_i
\end{equation}

$L_o$ is the outgoing radiance at surface point $\mathbf{x}$ in direction $\mathbf{\omega}_o$. 
The right hand side is an integral of the product of the incident radiance, $L_i(\mathbf{\omega}_i)$ (which includes a visibility term) from direction $\omega_i$ and the BSDF, $f_r(\mathbf{x}, \mathbf{\omega}_o, \mathbf{\omega}_i)$, which models the surface reflectance. The integration domain is the hemisphere, $\Omega$, around the surface normal$\mathbf{n}$.
Since this equation is infinitely recursive and hence intrectable, it is common practice to evaluate it with Monte Carlo estimators \cite{veach1998montecarlo, physically_based_rendering}, or simplify it with approximations like Split-Sum~\cite{karis2013real}.

We leverage a Cook--Torrance microfacet BRDF~\cite{cook1982reflectance},
\begin{equation}\label{eq:cook_torrance_specular_brdf}
f(\bomega_i,\bomega_o) \;=\;
   \frac{D\,G\,F}{4\,(\bomega_o\!\cdot\!\bn)\,(\bomega_i\!\cdot\!\bn)},
\end{equation}
whose specular term carries the reflection and highlight cues that drive the
Neural Renderer towards photorealistic relighting. For efficiency, we preintegrate the specular component with split-sum \cite{karis2013real,Munkberg_2022_CVPR}:
\begin{equation}\label{eq:splitsum}
L(\bomega_o)\;\approx\;
   \underbrace{\int_{\Omega} L_i(\bomega_i)\,D(\bomega_i,\bomega_o)\,(\bomega_i\!\cdot\!\bn)\,d\bomega_i}_{\text{prefiltered envmap}}\;\cdot\;
   \underbrace{\int_{\Omega} f(\bomega_i,\bomega_o)\,(\bomega_i\!\cdot\!\bn)\,d\bomega_i}_{\text{BRDF LUT}}.
\end{equation}
The first integral collapses to a roughness-mipped query of the envmap; the second is precomputed once into a 2D lookup table over
$(\bn\!\cdot\!\bomega_o,\,r)$. 

\section{Implementation Details}

\input{Figures/03_overall_pipeline}

\subsection{Prior-Informed Reconstruction.}
\label{sec:prior_informed_details}
Algorithm~\ref{alg:prior_informed_reconstruction} details the proposed two-stage prior-informed reconstruction procedure. For all reconstruction objectives, we employ the standard 3DGS loss function~\cite{kerbl3Dgaussians} (with $\lambda = 0.8$):
\begin{equation}
\label{eq:tron_recon_loss}
  \mathcal{L}_{\text{rec}}(I, \hat{I}) = \lambda \, \| I - \hat{I} \|_1 + (1-\lambda)\big(1 - \mathrm{SSIM}(I, \hat{I})\big).
\end{equation}
In Algorithm~\ref{alg:prior_informed_reconstruction}, $\mathcal{R}$ denotes the rendering operator defined according to the configurations listed in Table~\ref{tab:rendering_variants}.

Contrary to previous works, our setting avoids using additional regularizations like depth-distortion and normal-depth consistency. Following common practice \cite{kerbl3Dgaussians, 3dgrt2024}, we use 3rd order Spherical Harmonics to model baked radiance in the reconstruction. Similar to concurrent works \cite{poirierginter:hal-05306634}, we recognize that highly specular regions often suffer from degraded geometry, as very few views provide supervision for highlights. Still, rather than using lower order SH or replacing $\hat{I}_{rgb}$ with a non-directional diffusive prior, we found $\hat{I}_{rgb}$ to contain more crucial shade and texture cues to supervise the initial reconstruction.

Our prior prediction requirements are flexible. In this work we preprocess each view individually with an
intrinsic decomposition model model~\cite{DiffusionRenderer} for basecolor, metallic, roughness, normals and the model of \cite{zeng2024rgb} for irradiance. Flaws introduced during this preprocess phase are later mitigated when lifting to 3D and processing through our Neural Renderer. While priors can be extracted with video based versions of DiffusionRenderer~\cite{DiffusionRenderer} for improved view consistency, we found that comes at the cost of lower prior resolution, crucial to high frequency channels such as normals.

\subsection{Gaussian Scene Rendering} 
\label{sec:gs_details}
The rendering of an image pixel $\hat{I}_p$ for a camera with origin $\bo \in \mathbb{R}^3$ and viewing direction $\bd \in \mathbb{R}^3$ is defined by tracing a ray $r(\tau) = \bo + \tau \bd , \tau \geq 0$ and accumulating contributions from Gaussian particles along the ray in a sorted order:
\begin{equation}\label{eq:gs_rendering}
    \hat{I}_p = \sum_{j=1}^m \mathbf{f}_j(\bd)\; \alpha_j \prod_{k=1}^{j-1}(1-\alpha_k).
\end{equation}
Here $\alpha_j$ is the opacity of the $j^{\text{th}}$ depth-order Gaussian, computed from the maximal Gaussian kernel response along the ray and modulated by the learned particle opacity parameter $\sigma_j$. The function $\mathbf{f}_j(\mathbf{d})$ represents a general per-particle attribute to be rendered, potentially dependent on the viewing direction $\mathbf{d}$, such as view-dependent color, base color, surface normals, or other geometric or material properties. Table~\ref{tab:rendering_variants} summarizes the rendering quantities obtained by instantiating Eq.~\eqref{eq:gs_rendering}. Notably, our adoption of the 2D Gaussian formulation~\cite{Huang2DGS2024,gu2024IRGS} yields an explicit and well-defined per-particle \emph{geometric} normal, which we leverage throughout our method.

\paragraph{Deferred shading.}
We build upon 3DGRT~\cite{3dgrt2024}, and for efficiency we replace the forward integrator, which shades every particle independently with a deferred volumetric integrator~\cite{deering1988deferredshading}. The sorted slab traversal is reused to gather a per-pixel G-buffer (position, normal, basecolor, roughness, metallic) by rendering each material channel through Eq.\ref{eq:3dgrt_volumetric}~\cite{zhu_2025_gsror,liang2025gusir}. The
shading-point depth is taken from a median cutoff: we accumulate transmittance along the ray and adopt the depth of the first particle past which it falls below $0.5$. The shading normal defaults to the geometric normal. 
The geometric normal is derived directly from the rotation matrix $R$:
\begin{equation}
\mathbf{n}^{geo} = \mathbf{\tilde{n}}  \mathbf{R}^\top \quad \text{where } \mathbf{\tilde{n}} = \begin{bmatrix} 0 & 0 & 1 \end{bmatrix}^\top.    
\end{equation}

We evaluate the direct illumination integral using 
 split-sum \cite{karis2013real,Munkberg_2022_CVPR} as described in Eq.~\ref{eq:splitsum}.
The renderer produces linear HDR values, which are mapped to the unit range with AgX tonemapping~\cite{AgXtonemapper} before being passed to the Neural Renderer. We denote the resulting image $\hat{I}_{pbr}$.

The split-sum approximation does not model occlusions, so we complement $\hat{I}_{pbr}$ with a
ray-traced irradiance map, $\hat{I}_{irr}$, that integrates incident radiance
over the visible upper hemisphere. At each shading point we trace $K$ shadow
rays via multiple importance sampling~\cite{veach1998montecarlo}, combining cosine sampling
and light importance sampling using a piecewise-constant 2D distribution sampling technique~\cite{physically_based_rendering}, 
and query each ray for visibility against $\cG$. 
Visibility is the complement of transmittance, and transmittance along a ray is order-invariant in the
volumetric formulation~\cite{R3DG2023}, so we trace shadow rays in an out-of-order routine that avoids the per-hit insertion-sort otherwise
required for compositing (this is implemented with any-hit shader). When a ray escapes the field, we fetch the envmap by differentiable bilinear texture lookup.

\begin{table}[t]
\centering
\begin{tabular}{l c p{6cm}}
\toprule
\textbf{Rendered buffer} & \(\mathbf{f}_j(\mathbf{d})\) & \textbf{Description} \\
\midrule
RGB image 
& \(\mathrm{SH}(\mathbf{d}; \bbeta_j)\)
& View-dependent color represented using spherical harmonics coefficients \\

Base color 
& \(\bb_j \in \mathbb{R}^3\)
& Intrinsic diffuse (albedo) color \\

Roughness 
& \(r_j \in \mathbb{R}\)
& Surface roughness parameter \\

Metallicity 
& \(m_j \in \mathbb{R}\)
& Material metallicity parameter \\

Normal 
& \(\displaystyle 
\mathbf{n}_j = 
\bR_j \mathbf{e}_3

\)
& Per-particle geometric normal derived from Gaussian orientation \\
\bottomrule
\end{tabular}
\caption{Instantiation of Eq.~\eqref{eq:gs_rendering} for different rendered buffers. Each choice of \(\mathbf{f}_j(\mathbf{d})\) corresponds to rendering a different scene attribute.}
\label{tab:rendering_variants}
\end{table}

\include{Tables/a_algorithm1_prior_informed_reconstruction}
\include{Tables/b_algorithm2_envmap_optimization}

\subsection{Envmap Optimization} 
\label{sec:app:envmap_optimization}

In order to train the neural renderer, we require paired of inputs $(\hat{I}_{\mathrm{pbr}},\,\hat{I}_{\mathrm{irr}})$ and their matching high quality rgb image $I^{\star}$. For real world data, $I^{\star}$ is only available without $(\hat{I}_{\mathrm{pbr}},\,\hat{I}_{\mathrm{irr}})$ that match the lighting conditions of the rgb image. In order to implement the real data curation described in Section~\ref{sec:neural_renderer:training}, we apply the method described in Section~\ref{sec:tracing} in a differentiable manner. We apply the pipeline of \cite{zeng2024rgb} to obtain a pseudo-gt irradiance estimation $I_e$, and supervise against our own computation of irradiance based on reconstructed scene geometry, optimizing the intensity values of the envmap in the process. To better facilitate high dynamic range, envmaps are queried with an exponential activation. For this data generation step only, we also mask out the sky from $I_e$ and $\hat{I}_{\mathrm{irr}}$ using an off-the-shelf segmentation model \cite{carion2025sam3segmentconcepts}, as sky is emissive and Gaussian particles often exist in that region. The entire stage is described in Algorithm~\ref{alg:envmap_optimization}. Finally, in order to calibrate the envmap intensity scale, we run a small number of iterations to supervise a single scale parameter using rendered $\hat{I}_{\mathrm{pbr}}$ and $I^{\star}$.
Note that at inference time, we do not require the backwards pass and apply only the forward mode of Section~\ref{sec:tracing}.

\subsection{Hyperparameters, Hardware \& Training Time}

\input{Sections/06_implementation}

\section{Dataset Construction}
\label{sec:dataset_details}

\subsection{Data Pipeline}

\paragraph{\dldvbenchmark{} Details.} Our real-world evaluation set, \dldvbenchmark{}, consists of 33 scenes from the 2nd partition of DL3DV-10K~\cite{ling2024dl3dv}. These scenes were never used during training of the neural renderer. Each scene is paired with 3 different rotated environment maps from a curated collection of envmaps of different contrast levels, rotated to match the scene. The entire dataset includes 99 video clips of varying illuminations. For Gaussian reconstruction methods, we augment the dataset with SFM initialization provided by \cite{fcgs2025}.

\paragraph{\ourmethod{}-Synthetic Details} As full-scene multi-illumination datasets with known environment maps are scarce, we render out a synthetic dataset, closely following prior work~\cite{DiffusionRenderer}.
To construct \tronsynth{} Benchmark, we collect 29 high-quality 3D models from PolyHaven and a diverse set of 33 medium and high-contrast HDR environment maps from multiple sources \S\ref{sec:appendix_envmaps}. \tronsynth{} contains 15 single-object scenes and \tronsynth{} 25 multi-object scenes, each rendered with a path tracer under 5 sampled illumination conditions. Objects are arranged on a textured PBR floor with collision checks to ensure physically plausible layouts. For each scene, we sample cameras using two protocols:
    (1) \textbf{Spiral cameras} - a trajectory of 290 viewpoints, covering pole to grazing views, which is used by all reconstruction-based baselines and \ourmethod{} to reconstruct the original scene, and
    (2) \textbf{Oscillating video cameras} - smooth continuous camera trajectories of 93 frames, suited for both video model baselines and reconstruction-based baselines, thereof used by methods as test views.
All images in the benchmark were rendered with resolution of 800 x 800.

\subsection{Synthetic Benchmark Assets}
\label{sec:appendix_envmaps}

\subsubsection{High-Contrast Environment Maps}
Sourced from Poly Haven:
\begin{enumerate}
    \item \texttt{afrikaans\_church\_interior\_4k}
    \item \texttt{carpentry\_shop\_01\_4k}
    \item \texttt{construction\_yard\_4k}
    \item \texttt{ferndale\_studio\_01\_4k}
    \item \texttt{ferndale\_studio\_06\_4k}
    \item \texttt{future\_parking\_4k}
    \item \texttt{goegap\_road\_4k}
    \item \texttt{monochrome\_studio\_01\_4k}
    \item \texttt{moon\_lab\_4k}
    \item \texttt{qwantani\_noon\_puresky\_4k}
    \item \texttt{snowy\_field\_4k}
    \item \texttt{story\_studio\_05\_4k}
    \item \texttt{studio\_small\_01\_4k}
    \item \texttt{studio\_small\_07\_4k}
    \item \texttt{teutonic\_castle\_moat\_4k}
    \item \texttt{thatch\_chapel\_4k}
    \item \texttt{white\_home\_studio\_4k}
    \item \texttt{wooden\_lounge\_4k}
    \item \texttt{yellow\_field\_4k}
\end{enumerate}

\subsubsection{Medium-Contrast Environment Maps}
Sourced from HDRI-Hub:
\begin{enumerate}
    \item \texttt{container\_free\_Env}
    \item \texttt{Harbor\_3\_Free\_Env}
    \item \texttt{HDR\_040\_Field}
    \item \texttt{HDR\_041\_Path}
    \item \texttt{HDR\_110\_Tunnel\_Env}
    \item \texttt{HDR\_111\_Parking\_Lot\_2\_Env}
    \item \texttt{HDR\_112\_River\_Road\_2\_Env}
    \item \texttt{HDR\_Free\_City\_Night\_Lights\_Ref}
    \item \texttt{night\_free\_Env}
    \item \texttt{Stonewall\_Env}
\end{enumerate}

Sourced from ambientCG:
\begin{enumerate}
    \setcounter{enumi}{10} 
    \item \texttt{IndoorEnvironmentHDRI009\_4K\_HDR}
    \item \texttt{IndoorEnvironmentHDRI020\_4K\_HDR}
    \item \texttt{IndoorEnvironmentHDRI021\_4K\_HDR}
    \item \texttt{NightEnvironmentHDRI010\_4K\_HDR}
\end{enumerate}

\section{Evaluation Details}

\subsection{Inference Mode}

To evaluate \ourmethod{} on a new scene, we first obtain material priors per view (\S\ref{sec:prior_informed_rec}) and reconstruct the scene using our pipeline (\S\ref{sec:gs_rep}). 
Given a novel HDR environment map, we use our Gaussian Ray Tracer (\S\ref{sec:tracing}) to render $\hat{I}_{\text{pbr}}$ and $\hat{I}_{\text{irr}}$, which are then forwarded to the Neural Renderer (\S\ref{sec:neural_renderer}) to produce final RGB image. Given our extended Gaussian representation from Eq.~\eqref{eq:gaussian-representation}, editing becomes easy through recombination of particles from other scenes with positions of their subsets shifted, or material properties modified.

During interactive applications, we apply the neural renderer in image preview mode, operating on single frames without temporal consistency. Mirroring the iterative workflows of Blender~\cite{blender} and Cycles, this mode prioritizes immediate visual feedback over temporal consistency during scene drafting, while the complete framework is utilized for high fidelity finalization.

\subsection{Baseline Details}

\paragraph{Gaussian Methods.} Evaluation of normals is conducted in camera space. For methods that render normals in world space (GI-GS, GS-IR), we normalize and rotate the normals of each pixel using the test view cameras. For GaussianShader, we found the normals consistency loss on synthetic-benchmark causing severe artifacts due to the depth derived reference normal being too noisy. For this dataset-benchmark only, we reduce the normal consistency loss weight to zero. We repeated the experiments of all baselines using their latest code released, and validated against results reported in the respective papers. Discrepancies in reported numbers are due to code changes since paper publications, and different evaluation protocols.

\paragraph{Neural Rendering Methods.} For UniRelight and Diffusion Renderer, we project each envmap to camera space. As both methods do not require training data, for \dldvbenchmark{} we feed the entire trajectory consisting of train and test views, and select the test frames as output.

\subsection{Albedo Scaling}
\label{app:albedo_scaling}

For basecolor (albedo) metrics, we apply per-channel scaling channel optimization restricted to foreground pixels, a practice introduced by NeRFactor \cite{zhang2021NeRFactor} and adopted as standard by TensoIR \cite{Jin2023TensoIR} and subsequent Gaussian based inverse rendering works \cite{liang2023gs} and \cite{chen2025gigs}.
Concretely, for each test view we solve a closed-form least-squares problem to find scale $s$ that best aligns the albedo prediction of channel $c$ of view $i$ with the ground truth: $s_{i,c}^{\star} = \arg\min_{s \ge 0} \|s\,\hat{A}_{i,c} - A_{i,c}\|_2^2$. We repeat this convention for all baselines and datasets we evaluate on.

\subsection{A/B Photorealism Additional Details}

The A/B photorealism benchmark described in Tab.~\ref{tab:photorealism} was conducted using GPT 5.1. We collect 99 video clips by rendering \dldvbenchmark{} with each method. In each round we sample one result from ours, and one from the baseline. The agent is presented with a set of instructions, instructed to choose which anonymous image fits the instructions better: A, B or tie. Images are permuted, and pairs are randomized per trial. We repeat this test with 3 different seeds and report standard deviation within the table.
 
\label{app:agent_metrics}

\begin{tcolorbox}[
    enhanced,
    colback=gray!5, 
    colframe=blue!50!black, 
    title=Photorealism Evaluation Prompt,
    fonttitle=\bfseries\sffamily,
    arc=2mm,
    boxrule=0.8pt,
    width=\textwidth,
    drop shadow
]
\small
\noindent\textbf{Role:} You are evaluating two images, A and B, which are different rendering methods of the same 3D scene under the same lighting environment. Your task is to decide which looks more photorealistic.

\smallskip
\noindent\textbf{Setup:} Both images depict the same camera viewpoint and illumination (same HDR environment map). They differ only in the rendering method. Absolute image size is not a discriminator. Judge purely on rendering quality—not on content, framing, or composition.

\smallskip
\noindent\textbf{Criteria (in priority order):}
\begin{enumerate}[leftmargin=*, nosep]
    \item \textbf{Light transport:} Are specular highlights and reflections placed where geometry and lighting predict? Does indirect light look plausible?
    \item \textbf{Shadow consistency:} Are shadow positions, softness, and falloff coherent with light sources? Are contact shadows present?
    \item \textbf{Material plausibility:} Do surfaces read as the correct material? Avoid plastic-looking metals, chalky albedos, or oversaturated highlights.
    \item \textbf{Geometric coherence:} Are silhouettes clean? No popping, ghosting, or isolated artifacts indicating instability.
    \item \textbf{Color \& exposure:} Are tints coherent? Does brightness fall off naturally? Is the white balance plausible?
    \item \textbf{No neural artifacts:} Penalize blocky/checkerboard textures, hallucinated detail, smeared edges, or any "AI-look" that breaks physical plausibility.
    \item \textbf{Daylight plausibility:} 
    \begin{itemize}[label=--]
        \item Daytime scenes must \textbf{not} look uniformly dark. Globally dim renders where shadows should be present are penalized as failed rendering.
        \item Conversely, washed-out daylight lacking strong shadows where expected is also penalized.
    \end{itemize}
\end{enumerate}

\smallskip
\noindent\textbf{What to ignore:} Camera-level effects (lens distortion, vignette) and aesthetic preference. The goal is realism, not beauty.

\smallskip
\noindent\textbf{Anti-bias instruction:} Position (A or B) must not influence your decision. Imagine labels swapped, would your answer change?

\medskip
\noindent\colorbox{blue!10}{\textbf{Output:}} \textbf{A single token.} Use ``A'' or ``B''. Use ``TIE'' only when no meaningful difference exists.
\end{tcolorbox}

\include{Figures/appendix/agent_pairs}

Per-frame FPS and first-frame latency were measured on a single NVIDIA A100 for UniRelight\cite{he2025unirelight} and DiffusionRenderer\cite{DiffusionRenderer}. \ourmethod{} and Gaussian based methods \cite{chen2025gigs, liang2023gs, jiang2023gaussianshader} were measured on NVIDIA L40. We also report FPS of 1.6 and latency of 625ms for our pipeline running on a desktop machine equipped with NVIDIA A6000.

\subsection{VBench Metrics}
\label{app:content_only_metrics}
We apply the VBench \cite{zhang2024evaluationagent} protocol on 99 differently illuminated clips from \dldvbenchmark{}. Content-only metrics operate purely on pixels and cross-frame features, evaluating the generated output without requiring a ground-truth reference. We measure the following dimensions of the benchmark, which do not require a textual prompt.

\textbf{Motion Smoothness.} Measures temporal coherence with AMT-S optical flow. 

\textbf{Subject Consistency.} Evaluates whether the foreground subject remains coherent across the camera trajectory, using DINO across frames.

\textbf{Background Consistency.} Assesses whether background elements remain stable throughout the video, using CLIP across frames.

\textbf{Aesthetic Quality.} Estimates the overall perceived visual appeal of the frames using LAION over ViT-L/14. 

\textbf{Imaging Quality.} Measures per-frame image quality independent of any reference images.

%% file: Figures/fig_random_seed_comparison.tex
\begin{figure}[ht]
\centering
\includegraphics[width=\linewidth]{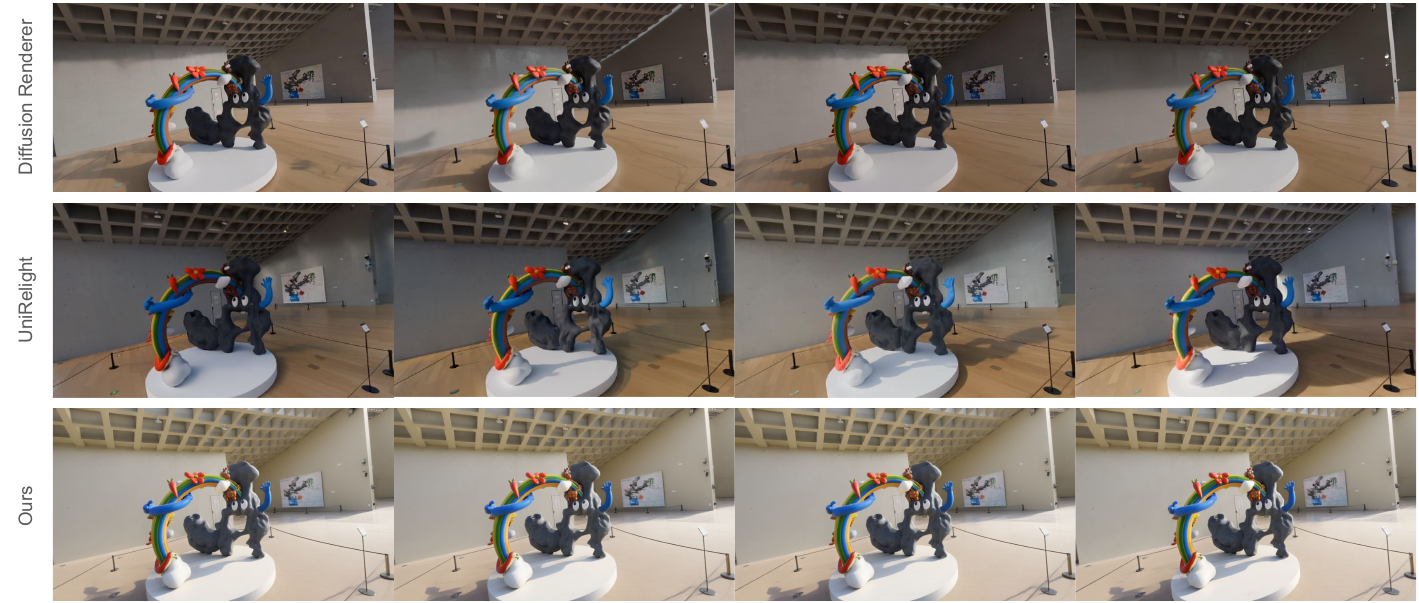}
\caption{
\textbf{Random seed sensitivity comparison}: 
We evaluate the sensitivity of each method to random initialization by running each model four times on the same input image with different random seeds. This test measures whether a model consistently follows the input conditioning or instead hallucinates lighting and shadow details. UniRelight~\cite{he2025unirelight} and Diffusion Renderer~\cite{DiffusionRenderer} are both multi-step diffusion models, making their outputs sensitive to the initial noise. In practice, we observe large variations across seeds: the cast shadow on the floor and the shadow on the background wall can change substantially in shape and appearance. This indicates that these methods perform a significant amount of generative reconstruction when producing shadows. By contrast, our method remains consistent across random seeds, owing to its single-step image-to-image design and strong input conditioning.
}\label{fig:randomseed}
\end{figure}

%% file: Figures/appendix/dl3dv_material_decomposition.tex
\begin{figure}[h]
    \centering
    \includegraphics[width=\textwidth]{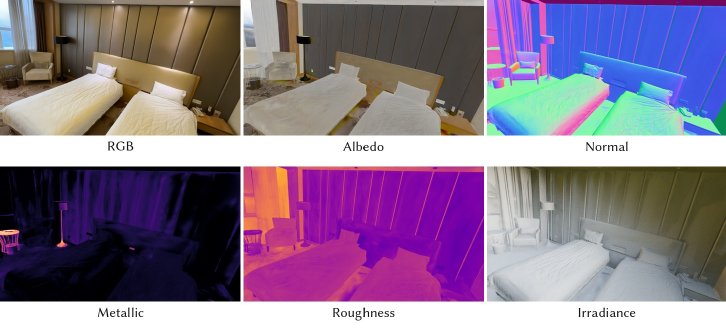}
    \caption{
        \textbf{Additional material decomposition results.} Showing gbuffer representation baked into the Gaussian field, each channel rendered individually.
    }
    \label{fig:dl3dv_material_decomp}
\end{figure}

%% file: Figures/appendix/relight-syntheticB-gauss.tex
\begin{figure}[h]
    \centering
    \includegraphics[width=\textwidth]{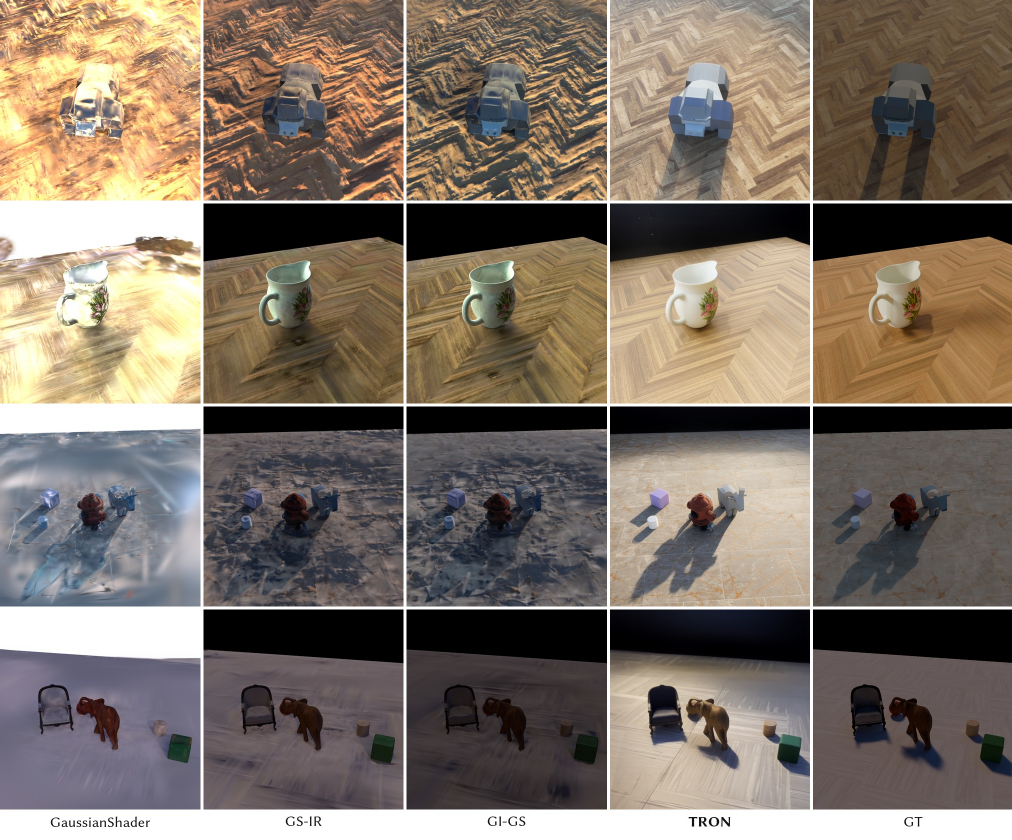}
    \caption{
        \textbf{Comparison to Gaussian based methods on the Relighting Synthetic Benchmark.} All methods were presented different illumination conditions during reconstruction. Top example: baselines exhibit baked shadows, in incorrect location and direction. Bottom row: long shadows due to novel relighting are not modeled by baselines, compared to \ourmethod{}.
    }
    \label{fig:relight-syntheticB-gauss}
\end{figure}

%% file: Figures/appendix/relight-syntheticB-neural.tex
\begin{figure}[h]
    \centering
    \includegraphics[width=\textwidth]{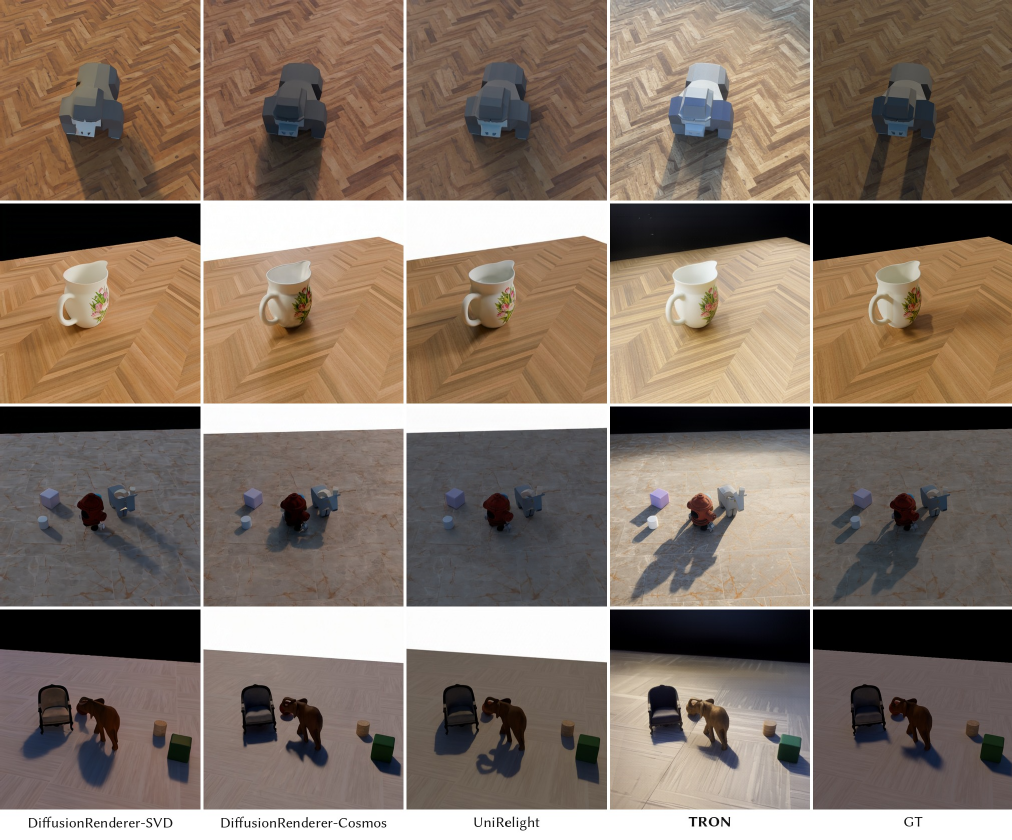}
    \caption{
        \textbf{Comparison to Neural based methods on the Relighting Synthetic Benchmark.} By adhering to irradiance cues rendered according to the geometry of the scene, \ourmethod{} models the correct silhouttes, faithful to the scene geometry. In comparison, other neural baselines condition on priors embedded in their weights, resulting in shadows that do not adhere to the intricate geometries of the scene.
    }
    \label{fig:relight-syntheticB-neural}
\end{figure}

%% file: Figures/appendix/mip360_gaussian_comparison.tex
\begin{figure}[h]
    \centering
    \includegraphics[width=\textwidth]{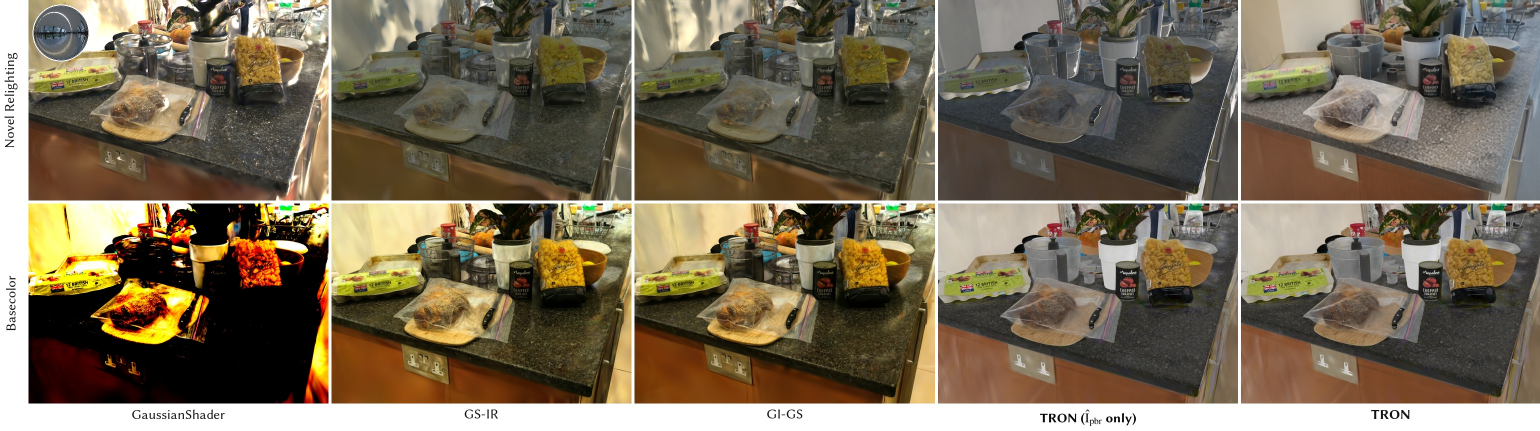}
    \caption{
        \textbf{Comparisons and Ablation of Tracer v.s. Full Pipeline on mip-360 \cite{barron2022mipnerf360}.} Left to right, columns 1-3: With shadows and highlights baked into the albedo, artifacts emerge during novel relighting. Column 4: Even with a clean albedo map, the shading model is limited in expressivity. Right column: the full pipeline benefits from the realism the neural renderer introduces.
    }
    \label{fig:mip360comp}
\end{figure}

%% file: Figures/fig_ablation_latent_grid.tex
\newcommand{\fusionGrid}[1]{%
  \begin{tikzpicture}
    \node[anchor=south west, inner sep=0] (img) at (0,0)
      {\includegraphics[width=0.92\linewidth]{#1}};
    \begin{scope}[
      shift={(img.south west)},
      x={(img.south east)}, y={(img.north west)}]
      \foreach \i/\w in {0/0.00, 1/0.25, 2/0.50, 3/0.75, 4/1.00} {%
        \pgfmathsetmacro{\cx}{(2*\i + 1)/10.0}%
        \node[above, yshift=2pt, font=\small, inner sep=1pt]
          at (\cx,1) {$w_{\mathrm{irr}}{=}\w$};
      }
      \foreach \j/\w in {0/0.00, 1/0.25, 2/0.50, 3/0.75, 4/1.00} {%
        \pgfmathsetmacro{\cy}{1 - (2*\j + 1)/10.0}%
        \node[left, xshift=-2pt, font=\small, inner sep=1pt]
          at (0,\cy) {$w_{\mathrm{pbr}}{=}\w$};
      }
    \end{scope}
  \end{tikzpicture}%
}

\newcommand{\guidanceInputs}[2]{%
  \begin{tabular}{@{}c@{\hspace{1.5em}}c@{}}
    \includegraphics[width=0.30\linewidth]{#1} &
    \includegraphics[width=0.30\linewidth]{#2} \\[2pt]
    {\small Guidance: $I_{\mathrm{pbr}}$ (PBR)} &
    {\small Guidance: $I_{\mathrm{img}}$ (irradiance)}
  \end{tabular}%
}

\begin{figure*}[p]
  \centering
  \guidanceInputs%
    {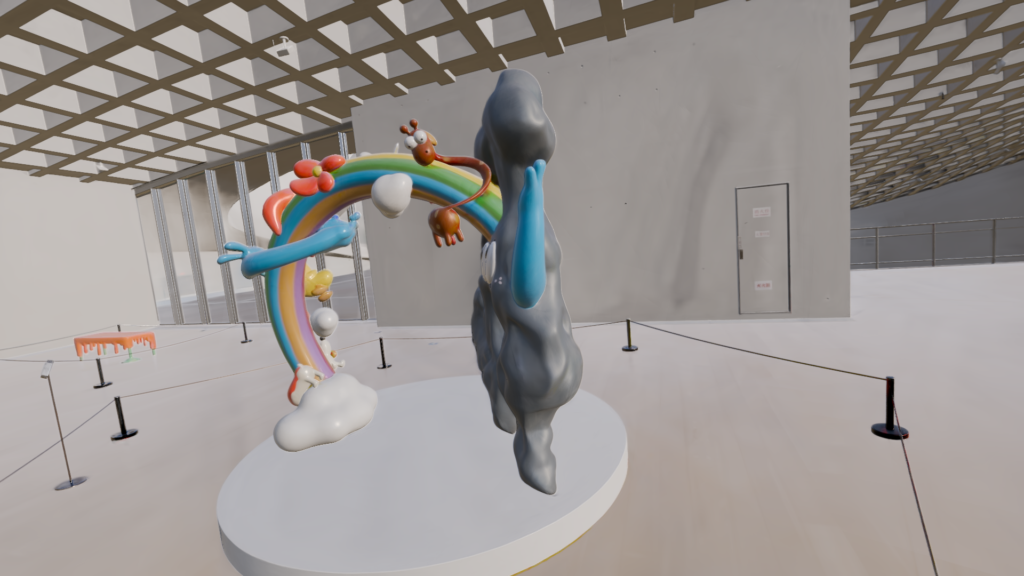}%
    {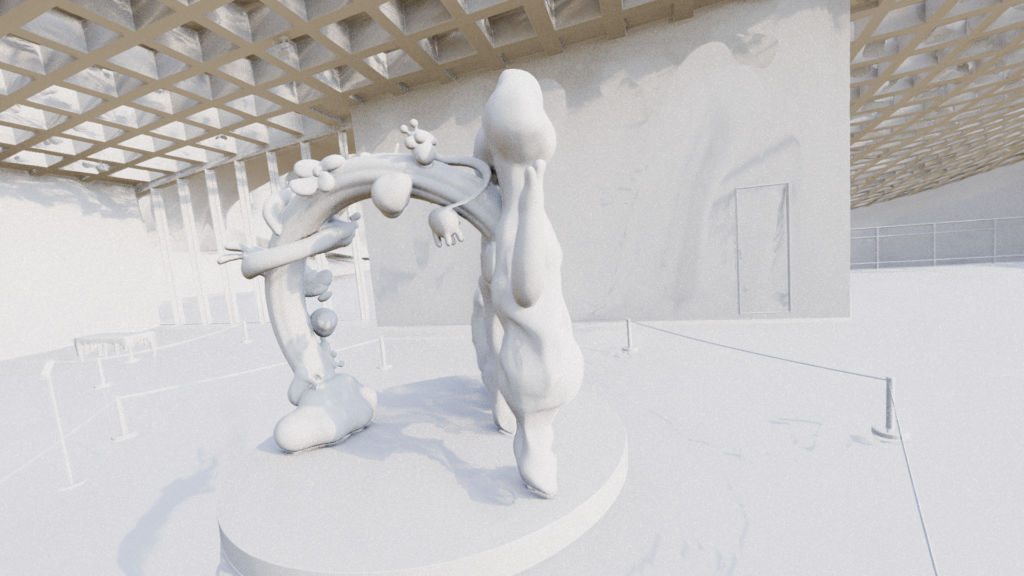}%
  \par\medskip
  \fusionGrid{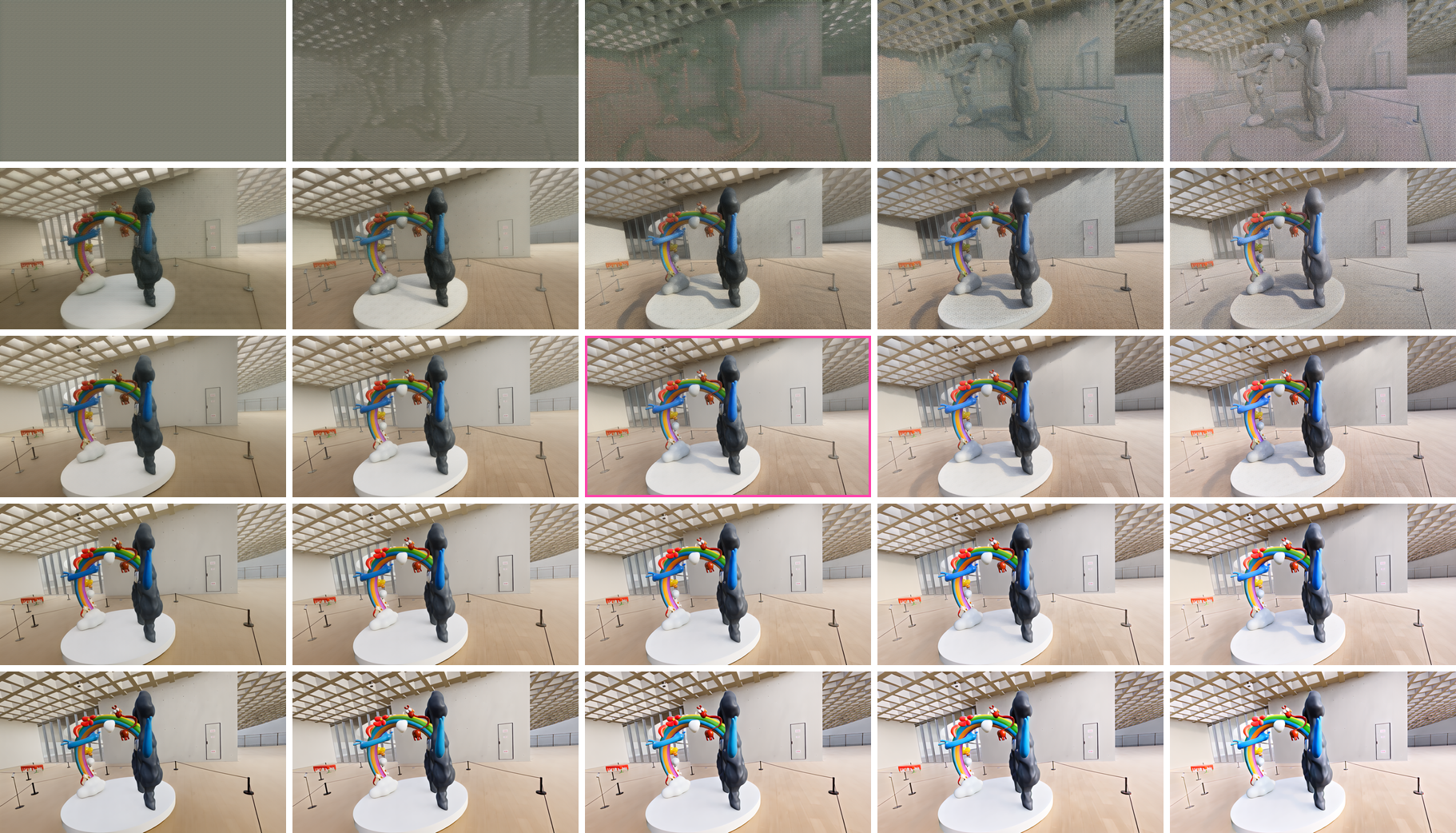}
  \caption{%
    \textbf{Latent-fusion weight sweep.}
    The Neural Renderer fuses the two input modalities by
    $I_{\mathrm{fused}} = w_{\mathrm{pbr}}\,I_{\mathrm{pbr}} + w_{\mathrm{irr}}\,I_{\mathrm{img}}$,
    where $I_{\mathrm{pbr}}$ and $I_{\mathrm{img}}$ are the standardized VAE
    latents of the rendered PBR and irradiance images shown at the top.
    The centre cell (outlined in magenta) is the default fusion settings
    $(w_{\mathrm{pbr}}, w_{\mathrm{irr}}) = (0.5, 0.5)$.
    The two axes carry distinct shading roles:
    increasing $w_{\mathrm{irr}}$ (left $\!\to\!$ right) progressively
    reintroduces cast shadows into the rendering, while increasing
    $w_{\mathrm{pbr}}$ (top $\!\to\!$ bottom) transitions from diffusive to specular highlights.
  }
  \label{fig:ablation_latent_weight_grid_2}
\end{figure*}

\begin{figure*}[p]
  \centering
  \guidanceInputs%
    {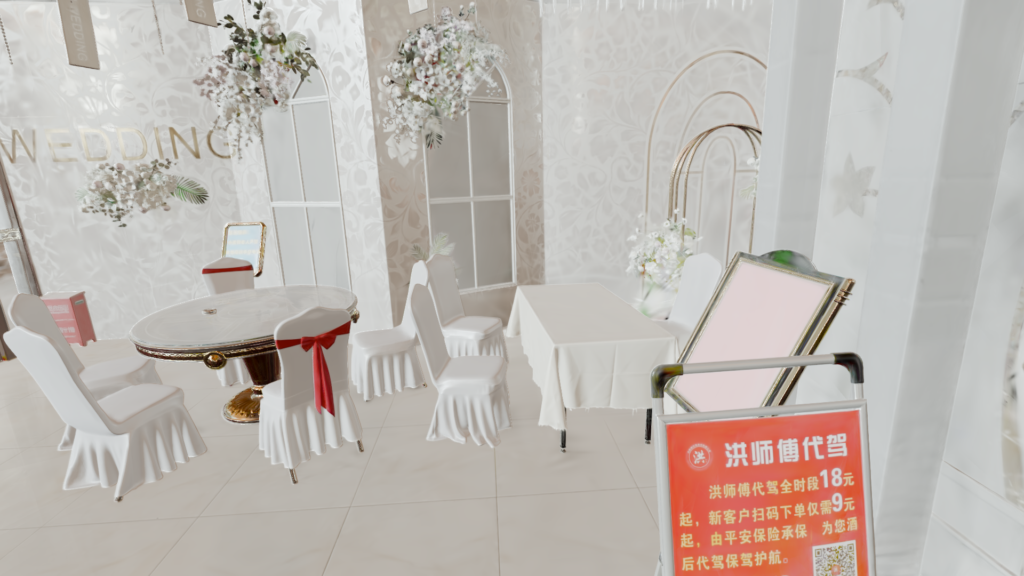}%
    {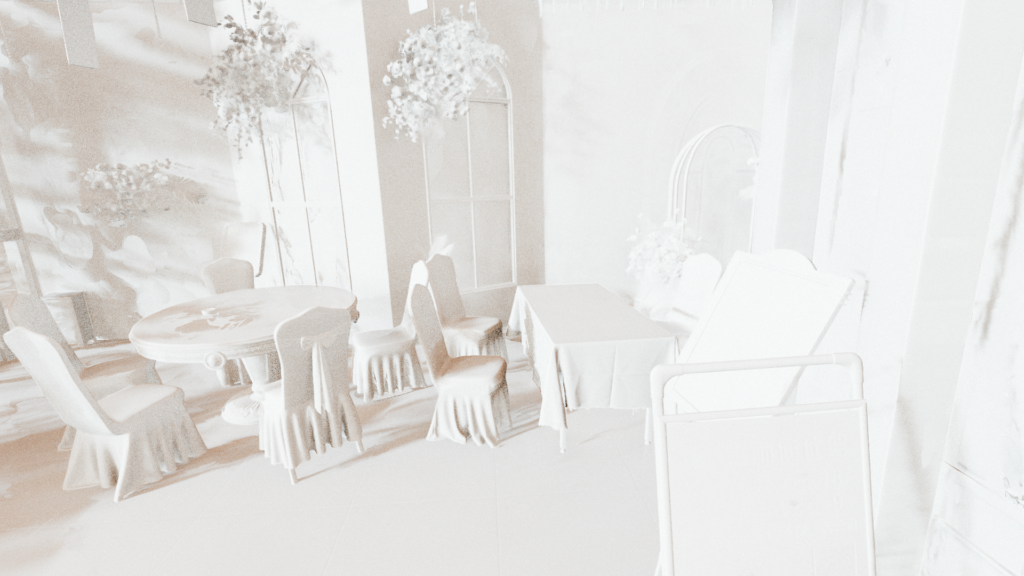}%
  \par\medskip
  \fusionGrid{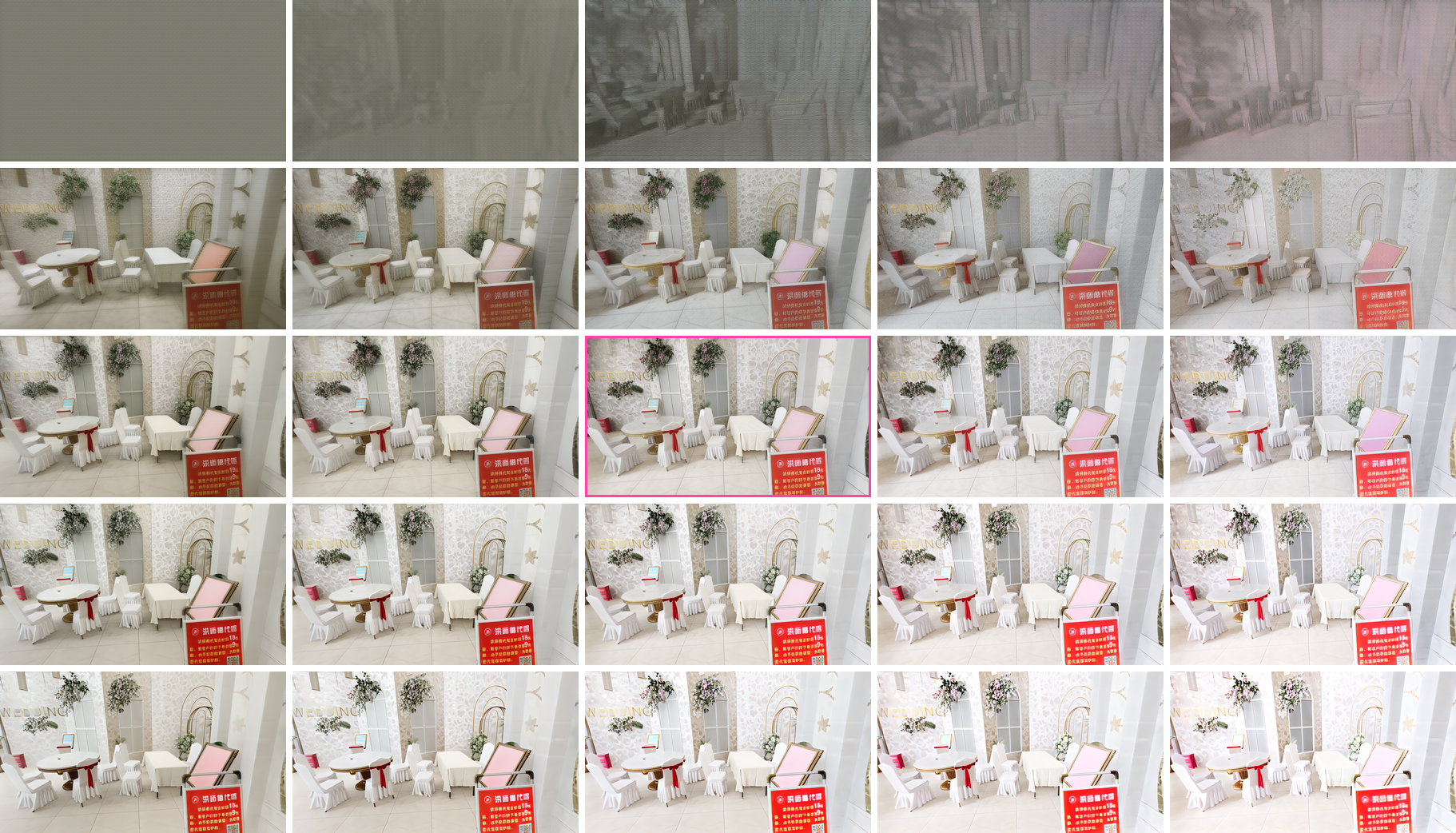}
  \caption{%
    \textbf{Additional latent-fusion weight sweep.}
    Same setup as Fig.~\ref{fig:ablation_latent_weight_grid_2}: rows vary
    $w_{\mathrm{pbr}}\!\in\![0,1]$ top to bottom, columns vary
    $w_{\mathrm{irr}}\!\in\![0,1]$ left to right, and the magenta-outlined
    centre cell is default fusion weights $(0.5,\,0.5)$. The two guidance
    inputs $I_{\mathrm{pbr}}$ and $I_{\mathrm{img}}$ are shown at the top.
    The same dichotomy holds: shadows naturally form with low $w_{\mathrm{irr}}$,
    while adhering closer to the guidance images as the irradiance weight increases.
  }
  \label{fig:ablation_latent_weight_grid_1}
\end{figure*}

%% file: Figures/03_overall_pipeline.tex
\begin{figure*}[t]
    \centering
    \includegraphics[width=\textwidth]{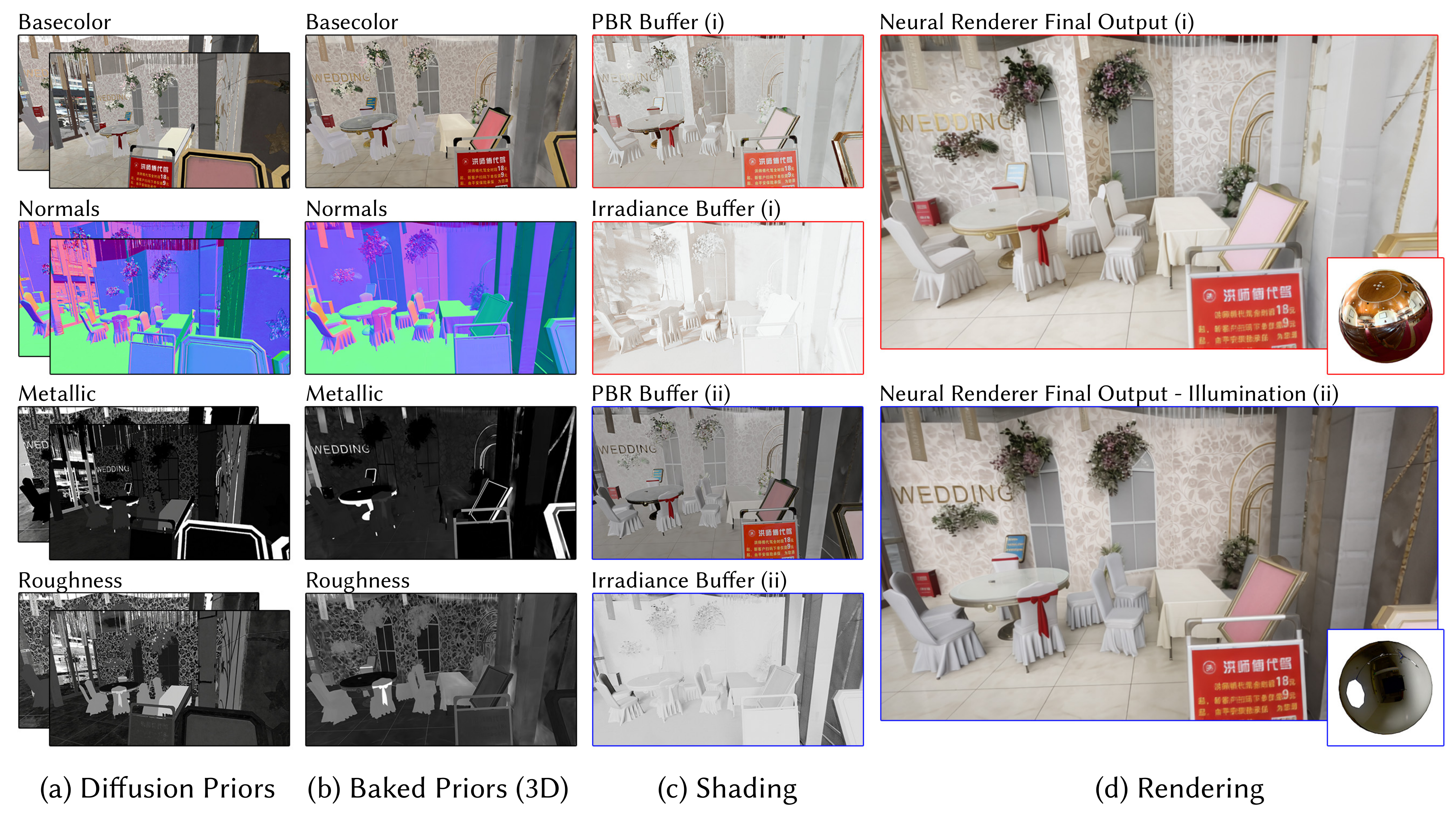}
    \caption{
        \textbf{Overall Pipeline:} (a) We extract per-frame material priors (b), and bake them into a 2D Gaussian representation as additional channels. (c) Given any novel lighting (envmap insets, rightmost), we can compute shading passes, PBR and irradiance (d), and use them to render a high-quality relit image with a Neural Renderer.
    }
    \label{fig:overall_pipeline}
\end{figure*}

%% file: Tables/a_algorithm1_prior_informed_reconstruction.tex
\begin{algorithm}[t]
\caption{Prior-Informed Gaussian Reconstruction}
\label{alg:prior_informed_reconstruction}
\begin{algorithmic}[1]

\Require Input images and cameras $\{(I_i, C_i)\}_{i=1}^N$, intrinsic decomposition model $\phi$
\Ensure Gaussian scene $\mathcal{G}=\{A_j;B_j\}_{j=1}^m$

\State \textbf{Intrinsic decomposition}
\For{$i=1,\dots,N$}
    \State $(I_n^i, I_b^i, I_r^i, I_m^i) \gets \phi(I_i)$
\EndFor

\Statex
\State \textbf{Stage 1: Baked appearance reconstruction}
\State Initialize Gaussian geometry and appearance parameters $\{A_j\}_{j=1}^m$

\For{$t=1,\dots,T_1$}
    \State Sample a training view $(I_i, C_i)$
    \State Render RGB image:
    \[
        \hat{I}_i \gets \mathcal{R}(\{A_j\}, C_i, \mathbf{f}^{\text{rgb}})
    \]
    \State Render normal map:
    \[
        \hat{I}_n^i \gets \mathcal{R}(\{A_j\}, C_i, \mathbf{f}^{\text{normal}})
    \]
    \State Compute loss:
    \[
        \mathcal{L}_{\text{stage1}}
        =
        \mathcal{L}_{\text{rec}}(\hat{I}_i, I_i)
        +
        \lambda_n \mathcal{L}_{\text{rec}}(\hat{I}_n^i, I_n^i)
    \]
    \State Update $\{A_j\}$ using gradient descent
\EndFor

\Statex
\State \textbf{Stage 2: Material optimization}
\State Freeze $\{A_j\}$
\State Initialize material parameters $\{B_j\}_{j=1}^m$

\For{$t=1,\dots,T_2$}
    \State Sample a training view $(I_i, C_i)$
    \State Render material channels:
    \[
        (\hat{I}_b^i, \hat{I}_r^i, \hat{I}_m^i)
        \gets
        \mathcal{R}(\{A_j\}, \{B_j\}, C_i, \mathbf{f}^{\text{mat}})
    \]
    \State Compute loss:
    \[
        \mathcal{L}_{\text{stage2}}
        =
        \mathcal{L}_{rec}(\hat{I}_b^i, I_b^i)
        +
        \mathcal{L}_{rec}(\hat{I}_r^i, I_r^i)
        +
        \mathcal{L}_{rec}(\hat{I}_m^i, I_m^i)
    \]
    \State Update $\{B_j\}$ using gradient descent
\EndFor

\Statex
\State \textbf{Output}
\State $\mathcal{G} \gets \{A_j; B_j\}_{j=1}^m$
\State \Return $\mathcal{G}$

\end{algorithmic}
\end{algorithm}

%% file: Tables/b_algorithm2_envmap_optimization.tex
\begin{algorithm}[t]
\caption{Stage 3: Envmap Optimization}
\label{alg:envmap_optimization}
\begin{algorithmic}[1]

\Require Input images $\{(I_i, C_i)\}_{i=1}^N$, pre-trained Gaussian scene $\mathcal{G}=\{A_j;B_j\}_{m=1}^M$, irradiance prior model $\psi$
\Ensure Optimized Environment Map $E$ and global scale $\sigma$

\Statex \textbf{// Pre-processing: Extract Irradiance Priors}
\For{$i=1,\dots,N$}
    \State $I_e^i \gets \psi(I_i)$ \Comment{Estimate pseudo-GT irradiance}
\EndFor

\Statex
\State \textbf{Stage 3a: Environment Map Refinement}
\State Freeze geometry $\{A_j\}$ and materials $\{B_j\}$; Initialize cubemap $E$
\For{$t=1,\dots,T_{3a}$}
    \State Sample view $(I_i, C_i)$
    \State Render G-buffer: $\{\mathbf{p}, \mathbf{n}, \mathbf{m}\} \gets \mathcal{R}_{\text{deferred}}(\mathcal{G}, C_i)$ 
    \State Sample $K$ directions $\{\omega_k\}_{k=1}^K \sim p(\omega)$ using MIS on $E$ and BRDF
    \State Compute visibility $v_k \gets \text{Trace}(\mathcal{G}, \mathbf{p}, \omega_k)$ \Comment{Any-hit occlusion test}
    \State $\hat{I}_{\text{irr}}^i = \frac{1}{K} \sum_{k=1}^K W_k^{\mathrm{MIS}}\frac{E(\omega_k) \cdot v_k \cdot (\mathbf{n} \cdot \omega_k)}{p(\omega_k)}$ \Comment{Monte Carlo Estimator}
    \State $\mathcal{L}_{\text{stage3a}} = \| \hat{I}_{\text{irr}}^i - I_e^i \|_1$
    \State $E \gets E - \eta \nabla_E \mathcal{L}_{\text{stage3a}}$
    \State \textbf{Update mipmap pyramid} $\mathcal{M}(E)$ \Comment{Synchronize for next iteration}
\EndFor

\Statex
\State \textbf{Stage 3b: Global Intensity Alignment}
\State Freeze $\{A_j, B_j, E\}$; Initialize $\sigma \gets 1.0$
\For{$t=1,\dots,T_{3b}$}
    \State Sample view $(I_i, C_i)$
    \State \textbf{Compute mipmap pyramid} $\mathcal{M}(\sigma \cdot E)$ \Comment{Reflect current intensity scale}
    \State Render specular component using split-sum:
    \[
        \hat{I}_{\text{spec}} \gets \mathcal{R}_{\text{split-sum}}(\mathcal{M}, \mathbf{n}, \mathbf{m}) \quad \text{where } \text{level } \ell \propto \text{roughness}
    \]
    \State Compute shadowed diffuse: $\hat{I}_{\text{diff}} \gets \text{MC\_Estimate}(\mathcal{M}, \mathbf{n}, \mathbf{m}, \text{vis}=v_k)$
    \State $\hat{I}_{\text{pbr}}^i = \hat{I}_{\text{diff}} + \hat{I}_{\text{spec}}$
    \State $\mathcal{L}_{\text{stage3b}} = \mathcal{L}_{\text{rec}}(\hat{I}_{\text{pbr}}^i, I_i)$
    \State $\sigma \gets \sigma - \eta \nabla_\sigma \mathcal{L}_{\text{stage3b}}$
\EndFor

\State \Return $E_{\text{final}} \gets \sigma \cdot E$

\end{algorithmic}
\end{algorithm}

%% file: Sections/06_implementation.tex
\label{sec:implementation}

\paragraph{Ray Tracer.}
The tracer part of the pipeline extends from Moenne-Loccoz et al.~\cite{3dgrt2024}.

We use the recent Slang~\cite{bangaru2023slang} based implementation to obtain auto-differentiable kernels for the deferred integrator computing irradiance and bsdf.
\ourmethod{} uses the 2D kernel shape of~\cite{3dgrt2024}, embedded in a quad for fast ray-particle intersections, which gives us a natural definition of a geometric normal. We modify the integrator to skip Gaussians we hit from the back side, as we found these to yield inconsistent surface approximations. 

For loss computation (Eq.~\ref{eq:tron_recon_loss}), we use a weight of $\lambda=0.2$ for SSIM and $1-\lambda=0.8$ for L1 reconstruction loss. For Tensor-IR and the synthetic-benchmark, we also apply a mask loss with $\lambda_o=0.2$. We train each scene with 30K steps for the radiance field reconstruction phase, and 7K additional steps for lifting gbuffers. For envmap optimization during data generation, we apply 4K steps to learn the envmap through the irradiance prediction. For fair comparison, we do not leverage the recent additions of Markov Chain Monte Carlo \cite{kheradmand20243d} and PPISP (Bilateral Grid Filtering) \cite{deutsch2026ppispphysicallyplausiblecompensationcontrol} which improve the reconstruction quality. When transitioning from volumetric radiance optimization to gbuffer optimization, we initialize Gaussian basecolor from the rgb. During shading, we use Multiple Importance Sampling with 2 spp for interactive preview, and 64 spp for high quality frames. Where learned envmaps are used, we fix them to resolution of $512 \times 512$, and use the logic of \cite{Munkberg_2022_CVPR} to construct differentiable mipmaps efficiently.

Unless stated otherwise, our remaining hyperparameters follow the default configuration of \cite{3dgrt2024}.

\textbf{Neural Renderer.} The neural renderer draws from the architecture of \cite{zhang2026diffusionharmonizer}, using Cosmos-Predict-0.6B as backbone. We train the model with an AdamW optimizer and learning rate of $2e^{-5}$, in 3-phases of 32K iteration each, as described in Section~\ref{sec:neural_renderer}. 
In phase 1 of the training curriculum (Sec.\ref{sec:neural_renderer:training}), we train the model on synthetic data of resolution $512 \times 512$. For phase 2 and 3 we resize input frames to $1024 \times 576$.

For the entire length of optimization, we keep the WAN 2.1 VAE frozen and train only the DiT throughout the entire training. Different to \cite{zhang2026diffusionharmonizer}, our architecture does not assume skip connections between the encoder and decoder due to domain differences in input and output. The single-step denoising setting follows \cite{zhang2026diffusionharmonizer} and is prefixed to $t=250$. In all phases, we train the model with L2 and LPIPS \cite{lpips} loss, weighted equally. After encoding each pair of inputs with the VAE, latent fusion is achieved by weighted average, where out settings are fixed to $w_{pbr}=0.5$ and $w_{irr}=0.5$. In our full pipeline, the temporal dimension compresses 5 frames to a latent of 2-dim.

For interactive preview, we feed single $\hat{I}_{pbr}$ and $\hat{I}_{irr}$ frames, encoded without temporal consistency. Our final results are conducted with temporal compression of 5 frames.

\paragraph{Hardware.} Interactive applications which require a GUI were conducted on a machine equipped with an NVIDIA A6000 gpu. All comparisons reported in this paper, using the \ourmethod{} and Gaussian baselines \cite{jiang2023gaussianshader, liang2023gs, chen2025gigs} were conducted on a machine with an NVIDIA L40 gpu. For neural methods \cite{DiffusionRenderer, he2025unirelight}, we used a single A100 gpu.

Data generation pipeline was run on a cluster of 22 x NVIDIA L40 gpus. The reconstruction and rendering of 955 scenes from DL3DV-10K \cite{ling2024dl3dv} took 4 days end-to-end, including envmap optimization. The synthetic data generation, reconstruction and re-rendering lasted 72 hours. Each phase of the neural renderer was trained for 12 hours on a machine equipped with 8 x NVIDIA A100 gpus.

%% file: Figures/appendix/agent_pairs.tex
\begin{figure*}[t]
    \centering
    \includegraphics[width=\textwidth]{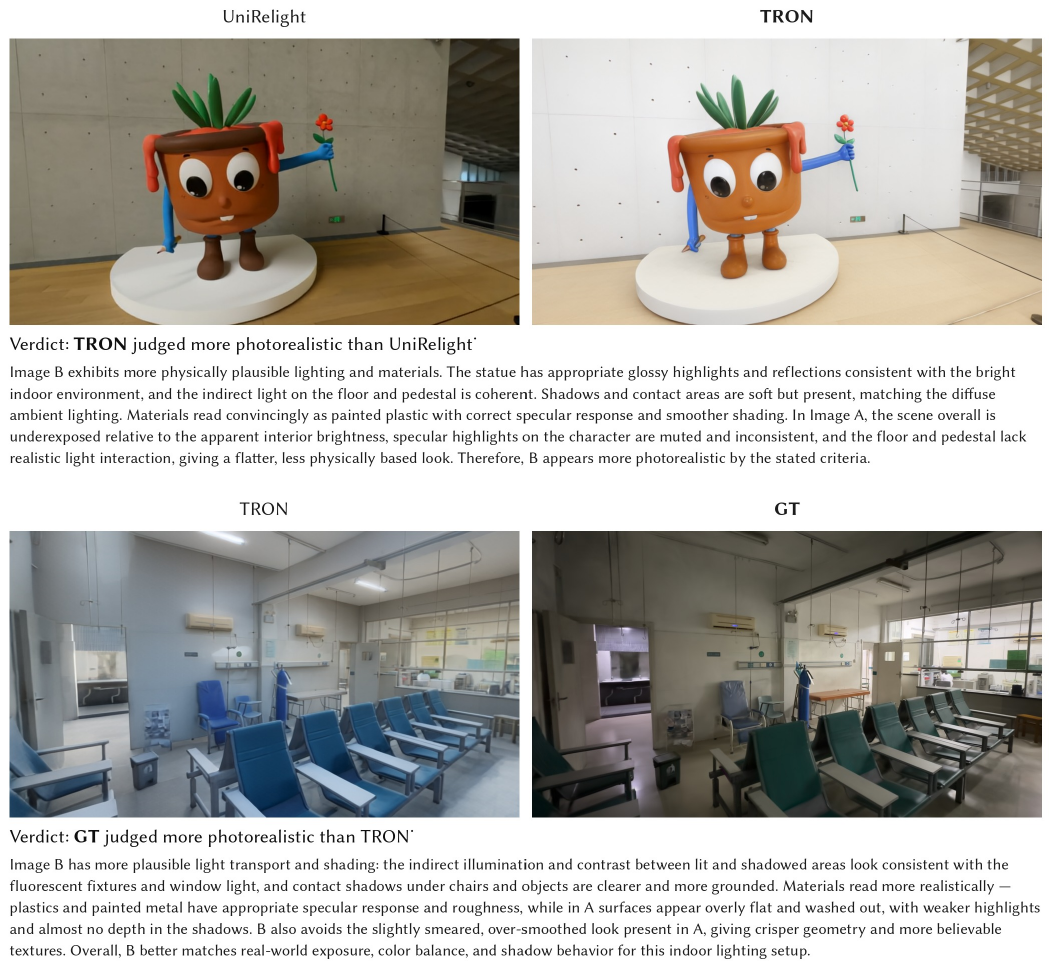}
    \caption{
        \textbf{Agent Test Explainability.} Showing rationale for final verdict of agent score, as presented in A/B photorealism benchmark of Table~\ref{tab:photorealism}. The proposed evaluation can be further validated with descriptions.
    }
    \label{fig:agent_pairs}
\end{figure*}